\documentclass[twoside,leqno,twocolumn]{article}

\usepackage[letterpaper]{geometry}
\usepackage{xurl}
\usepackage{ltexpprt}
\usepackage{hyperref}
\usepackage{epstopdf,xspace}
\usepackage{diagbox} 
\usepackage{bbding}

\usepackage[font={small,it}]{caption}
\usepackage{soul}
\usepackage{caption}
\usepackage{subcaption}
\usepackage{graphicx}
\usepackage{empheq}
\usepackage[algo2e,ruled,vlined, linesnumbered]{algorithm2e}
\usepackage[noend]{algpseudocode}
\usepackage{multirow}
\usepackage{tabularray}
\usepackage{adjustbox}
\usepackage{xcolor}

\usepackage{amssymb}

\DeclareMathAlphabet\mathbfcal{OMS}{cmsy}{b}{n}

\newcommand{\hide}[1]{}

\newcommand{\myTag}[1]{ {\bf \underline{#1}}\xspace} % for caption summaries
\newcommand{\myRot}[1]{ \rotatebox{90}{#1} } % for salesman matrix
\newcommand{\myck}{ \CheckmarkBold } % for salesman matrix

\newcommand{\scalable}{{Scalable}\xspace}
\newcommand{\Explainable}{{Explainable}\xspace}
\newcommand{\explainable}{{explainable}\xspace}

\newcommand{\domainSpec}{{domain-informative}\xspace}
\newcommand{\effective}{{effective}\xspace}

\newcommand{\Effective}{{Effective}\xspace}

\newcommand{\method}{EBV\xspace}

\newcommand{\methodTwo}{{EBV\textsubscript{model}}\xspace}

\newcommand{\myAlgorithm}{{EBV\textsubscript{fit\&cut}}\xspace}

\newcommand{\strengthHeating}{{s_h} \xspace}
\newcommand{\strengthCooling}{{s_c} \xspace}

\newcommand{\AllthetaExt}{{\theta_E} \xspace}
\newcommand{\AllthetaPeri}{{\theta_P} \xspace}
\newcommand{\AllthetaCore}{{\theta} \xspace}
\newcommand{\CutPoint}{{CP} \xspace}

\newcommand{\thetaCore}{{\Theta(t)} \xspace}
\newcommand{\thetaPeri}{{\Theta_{peri}(t)} \xspace}
\newcommand{\thetaExt}{{\Theta_{ext}(t)} \xspace}
\newcommand{\thetaAdj}{{\Theta_{adj}(t)} \xspace}
\newcommand{\thetaIdeal}{{\theta_{ideal}} \xspace}

\newcommand{\thetaCoreNorm}{{\theta(t)} \xspace}
\newcommand{\thetaPeriNorm}{{\theta_{peri}(t)} \xspace}
\newcommand{\thetaExtNorm}{{\theta_{ext}(t)} \xspace}
\newcommand{\thetaAdjNorm}{{\theta_{adj}(t)} \xspace}

\newcommand{\Principled}{{Principled}\xspace}

\newcommand{\explainability}{{Explainability}\xspace}
\newcommand{\effectiveness}{{Effectiveness}\xspace}
\newcommand{\scalability}{{Scalability}\xspace}
\newcommand{\observations}{{Observations}\xspace}

\newcommand{\thetaCoreFit}{{\hat{\Theta}(t)}\xspace}

\newtheorem{observation}{Observation}

\usepackage{enumerate}
\usepackage[shortlabels]{enumitem}
\setlist{leftmargin=5mm}

\newenvironment{tight_itemize}
{ \begin{itemize}
    \setlength{\itemsep}{0pt}
    \setlength{\parskip}{0pt}
    \setlength{\parsep}{0pt}     }
{ \end{itemize}                  } 

\newenvironment{tight_enumerate}
{ \begin{enumerate}
    \setlength{\itemsep}{0pt}
    \setlength{\parskip}{0pt}
    \setlength{\parsep}{0pt}     }
{ \end{enumerate}                  } 
\begin{document}

\title{\Large \method: Electronic Bee-Veterinarian for Principled Mining and Forecasting of Honeybee Time Series}
\author{Mst. Shamima Hossain\thanks{Dept.\ of Computer Science and Engineering, University of California, Riverside. Emails: \{mhoss037,vtsotras\}@ucr.edu.}
\and Christos Faloutsos\thanks{School of Computer Science, Carnegie Mellon University. Email: christos@cs.cmu.edu.}
\and Boris Baer\thanks{Dept. of Entomology, University of California, Riverside. Email: boris.baer@ucr.edu.}
\and Hyoseung Kim\thanks{Dept. of Electrical and Computer Engineering, University of California, Riverside. Email: hyoseung@ucr.edu.}
\and Vassilis J. Tsotras\footnotemark[1]}

\date{}

\maketitle

% Copyright Statement
% When submitting your final paper to a SIAM proceedings, it is requested that you include
% the appropriate copyright in the footer of the paper.  The copyright added should be
% consistent with the copyright selected on the copyright form submitted with the paper.
% Please note that "20XX" should be changed to the year of the meeting.

% Default Copyright Statement
\fancyfoot[R]{\scriptsize{Copyright \textcopyright\ 2024 by SIAM\\
Unauthorized reproduction of this article is prohibited}}

% Depending on which copyright you agree to when you sign the copyright form, the copyright
% can be changed to one of the following after commenting out the default copyright statement
% above.

%\fancyfoot[R]{\scriptsize{Copyright \textcopyright\ 20XX\\
%Copyright for this paper is retained by authors}}

%\fancyfoot[R]{\scriptsize{Copyright \textcopyright\ 20XX\\
%Copyright retained by principal author's organization}}

%\pagenumbering{arabic}
%\setcounter{page}{1}%Leave this line commented out.

\begin{abstract}
Honeybees are vital for pollination and 
food production.
Among many factors, extreme  temperature (e.g., due to climate change) 
is particularly dangerous for bee health.
Anticipating such extremities would allow beekeepers to take early preventive action.
Thus,
given sensor (temperature) time series data from beehives,
how can we find patterns and do forecasting?
Forecasting is crucial as it helps spot
unexpected behavior and thus issue warnings to the beekeepers.
In that case, what are the right models for forecasting?
ARIMA, RNNs, or something else?

We propose the \method (Electronic Bee-Veterinarian) method, which has the following desirable properties:
(i) {\em principled}: it is based on a) diffusion equations from physics and b) 
control theory for feedback-loop controllers;
(ii) {\em effective}: it works well on multiple, real-world time sequences,
(iii) {\em explainable}: it needs only a handful of parameters (e.g., bee strength)
that beekeepers can easily understand and trust,
and
(iv) {\em scalable}: it performs linearly in time.
We applied our method to multiple real-world time sequences,
and found that it yields accurate forecasting
(up to {\em 49\%} improvement in 
RMSE compared to baselines),
and segmentation. Specifically, 
discontinuities detected by \method mostly coincide with domain expert's opinions,
showcasing our approach's potential and practical feasibility.
Moreover, \method is scalable and fast, taking
about {\em 20 minutes} on a stock laptop for reconstructing two months of sensor data.
\end{abstract}


\section{Motivation \& Background}\label{sec:intro}
%How can we help beekeeping practitioners and industry to effectively monitor the health of beehives? 

The key problem we address is the design
of better tools so that beekeepers can effectively monitor the health of their hives
and take preventive action.
% and get early warning signs of environmental stressors that threaten their hives.

\smallskip\noindent{\bf Motivation.}
{\em Without honeybees, humanity will suffer.} Honeybees are vital pollinators. Their pollination services are required for the production of more than 80 crops or about a third of what we eat %--- either through direct consumption of fruits and vegetables or indirect consumption of crops by livestock 
\cite{eilers}. The yearly value of bee pollination has been estimated to be up to \$550 billion globally and \$29 billion for the U.S. alone \cite{potts2016safeguarding}. 
%Crops pollinated by bees are the source of 74\% of lipids via plant oil, 98\% of the world's Vitamin C, and more than 70\% of the world's Vitamin A \cite{eilers}. 
However, beekeepers report that they have lost 40\% of their colonies recently, e.g., a substantially higher colony loss of 45.5\% from April 2020 to April 2021 in the U.S.~\cite{aurellunited, forister2019declines}.
%sanchez2021further, sanchez2019worldwide
The reasons for the decline are multiple, including parasites, pesticides, and, most recently, climate change.

\smallskip\noindent{\bf Background.}
In the 2nd CIBER Bee Health Conference at UC Riverside~\cite{bee_conference}, practitioners were very emphatic on the need for a reliable and explainable hive monitoring system to help them understand the hive's internal condition for timely action. Thus, we need to stay away from black-box methods and instead use parsimonious, explainable models; such methods have few parameters, and they are easy to provide justification to domain experts for anticipated extremities.

Like a human body, hive temperature gives valuable information about bees' health. Bees tightly regulate their hive temperature (homeostasis) by heating during cold periods and cooling during the heat. This is known as hive \textit{thermoregulation}. 
The end result is that the hive temperature ranges between $33 - 36^\circ$C \cite{Kleinhenz, Petz2004RespirationOI}. Environmental stressors (e.g., parasites, etc.) can cause bees to lose their ability to maintain the hive's homeostasis. Thus, increased fluctuation of hive temperature can be seen as a first-order response indicating the start of a collapse.  %In the absence of insect pollinators, the use of manual-labor pollination is expensive and not sustainable. Thus, it has become vital to monitor honeybee colony health and alert the beekeepers for timely actions.% and ensure autonomous control for a healthy hive environment.  

\begin{figure*}
	\centering
	\subfloat[]{
		\includegraphics[height =3.9 cm]{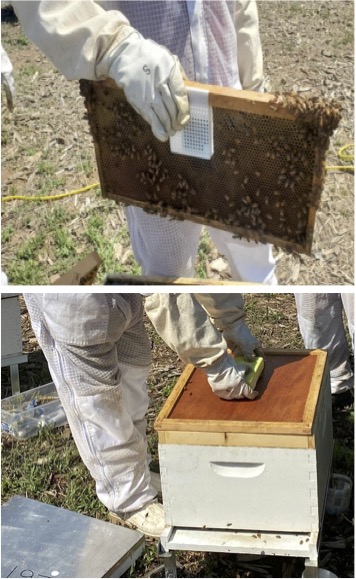}
	}
	\subfloat[]{
		\includegraphics[height =3.9 cm]{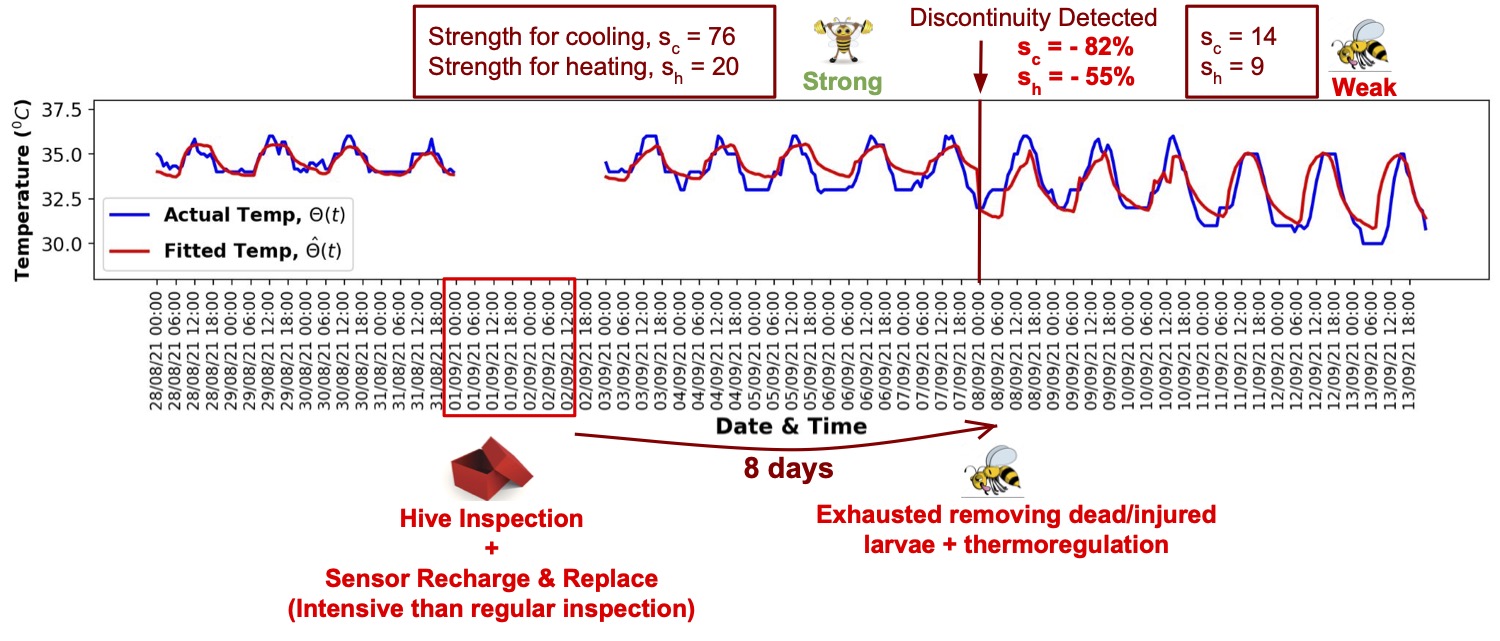}
	}
        \subfloat[]{
		\includegraphics[height =3.9 cm]{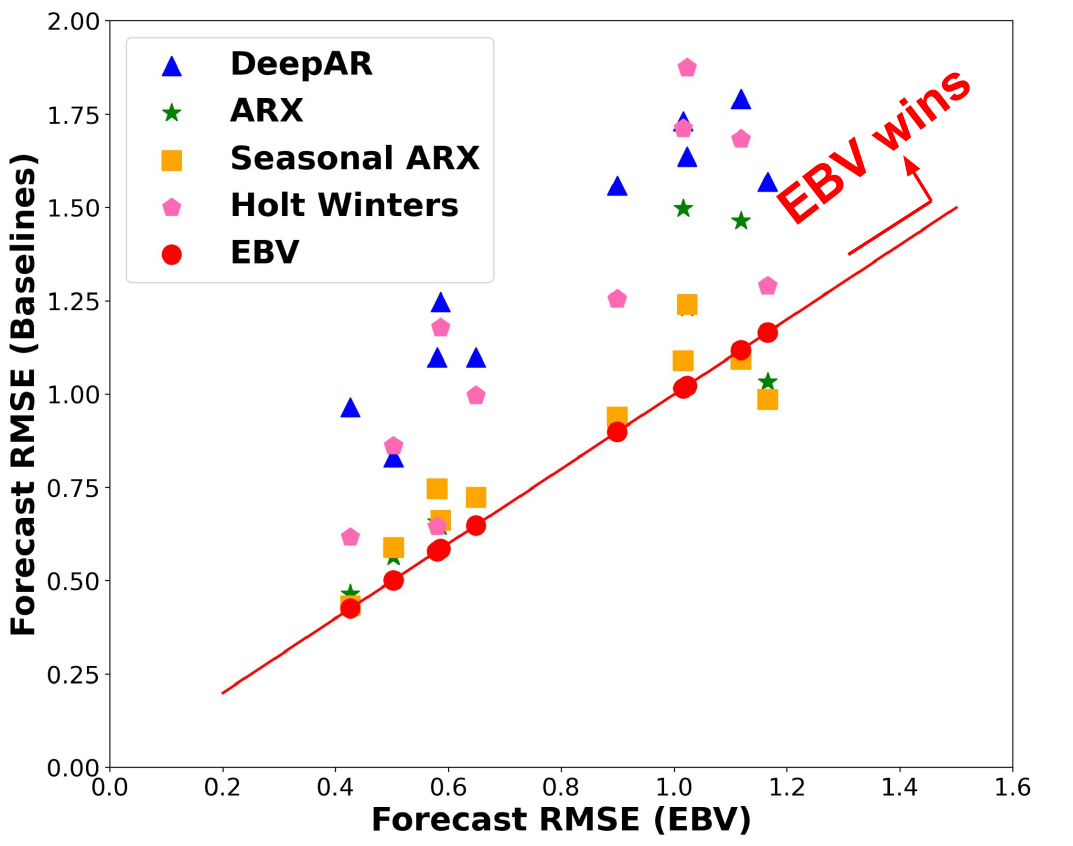}
	}\vspace{-1pt}	
	\caption{\myTag{\method at work:} 
        (a) deployed sensors
        (b) reconstruction and event detection
        (c) \method wins on forecasting.
        %Installing sensors onto the hive: (a) On the outer frame (Periphery area); (b) On the center frame (Core area).
        }\vspace{-1pt}	
	\label{fig:work_flow}	
\end{figure*}

%Climate change and rising environmental temperatures are emerging as the primary causes of hive collapse, higher mite infestations, and reduction in pollination \cite{Vercelli}. 

\smallskip\noindent{\bf Method overview.}
% To address the need of the beekeeping industry for functional bee health technologies, 
We propose a method, \myTag{\method} (Electronic Bee Veterinarian), specifically designed to monitor the hive temperature. We applied \method to multiple real-time sequences collected from different hive settings
%control and treated hives (treatment for combating heat stress during summer in California, USA) 
and achieved reliable forecasts for several days ahead 
%with an average error (RMSE) of 2.58 
with a good fit
(Figure \ref{fig:work_flow}(c), Details in Section \ref{sec: forecasting}), a lift of up to {\em 49\% } in AUC over baselines (Figure \ref{fig:accuracy});
computation time of around 20 minutes for fitting two months of data, and segmentation coinciding with external events affecting bees' thermoregulation ability (Figure \ref{fig:work_flow}(b), Details in Section \ref{sec:eval_segmentation}). %As \method does not depend on the size, type, or condition of the hives, we conjecture that it will be applicable to bees and bees-like insects maintaining thermoregulation. 

\smallskip\noindent\textbf{Contributions.} %The main contribution of this paper is the \method method, which is comprised of a mathematical model and an algorithm explaining the pattern of the hive core temperature pattern. 
Specifically, the main contributions of this paper are summarized as follows:
%Our \methodTwo method has the following properties:
\begin{tight_itemize}
    \item {\bf \Principled:} We propose a method, \method comprised of (i) \methodTwo, a novel model based on the thermal diffusion equation; control theory;
    and our novel `{\em split}' P-controller idea (Section \ref{sec:proposed_model}); and (ii) \myAlgorithm, a segmentation algorithm (Section \ref{sec: algorithm}) based on \methodTwo to estimate the number of cut-points where each cut represents a different hive condition. %\red{penalizing model complexity}. 
    As bees follow the same mechanism to regulate hive temperature, our method is universally applicable regardless of region, season, and bee species.

    \item {\bf \Effective:} We use the proposed \myAlgorithm for (i) fitting, (ii) forecasting, and (iii) segmentation to detect unexpected events. It performs well in terms of forecasting accuracy.
    
    \item {\bf \Explainable:} We can explain the current hive condition with the parameters (i.e., beehive strengths) measured and discontinuity detected by \myAlgorithm. %\myAlgorithm is able to explain the detected discontinuity affecting hive health.

    \item {\bf \scalable:} Our method is scalable and hands-off, requiring no parameter tuning. It is applicable to all hives irrespective of size and population.

    \item {\bf Informative:} Our experimental findings coincide with domain experts' expectations: (i) bees in treated hives are less affected by intensive hive inspection; (ii) treated colonies have a higher strength to maintain thermoregulation %(less fluctuation) 
    than control ones, and (iii) heating a hive is easier than cooling it down ~\cite{Jarimi2020ARO}.
    %and (iv) broods and larvae are the motivation for bees to thermoregulate \cite{Anton}. 

    %need to change observation
    % \end{itemize}
\end{tight_itemize}
\urlstyle{tt}
\smallskip\noindent\textbf{Reproducibility.} Our implementation is open-sourced at 
\url{https://github.com/rtenlab/EBeeVet}.
The dataset is available upon request. \footnote{Contact the authors or the CIBER bee research center at UC Riverside (\url{https://ciber.ucr.edu/}).}

\section{Method \& Technical Solution}

\subsection{Related Work} \label{sec:related_work}
\noindent\textbf{Beehive monitoring and analysis.} 
% Many systems have been proposed to collect sensor data from inside and
% outside the beehive. These systems 
Past systems collect in-hive microclimate parameters like temperature, humidity, and CO\textsubscript{2} ~\cite{Cecchi, Kridi2016ApplicationOW, pandimurugan2021iot, songa2022development}, sound signals ~\cite{Murphy2015AnAW, Zhange2015}, and visual bee activities using IR
LEDs or cameras installed at the hive entrance ~\cite{jiang2016wsn, Yang2017}.
Efforts have been made mostly to analyze sound signals to detect swarming events ~\cite{Murphy2015AnAW, Zhange2015}. Attempts to analyze other in-hive %sensor 
data such as temperature to detect anomalies ~\cite{padraig, Murphy2016bWSNSB} are limited to {\em arbitrary thresholds} defined by domain specialist, rule-based, or simple autoencoders. All these approaches fail to provide causal relationships with
colony health, lack robust evidence for analysis, and suffer from
poor sensing quality or intermittent interruptions due to environmental conditions. 

\begin{table}[h]
    \begin{center}
    \small
        \begin{tabular}{l | c| c  | c || c}
            \diagbox{Property}{Method} 
                %& \myRot{Prior hive mon.~\cite{padraig, Murphy2016bWSNSB}} 
                & \begin{tabular}[c]{@{}c@{}}Prior\\hive\\mon.\\\cite{padraig, Murphy2016bWSNSB}\end{tabular}
                 %& \myRot{Trad. time series ~\cite{Box15, Hamilton}} 
                 & \begin{tabular}[c]{@{}c@{}}Trad.\\time\\series\\\cite{Box15, Hamilton}\end{tabular}
                 %& \myRot{Deep Learning Tech. ~\cite{Autogluon}}
                 & \begin{tabular}[c]{@{}c@{}}Deep\\Learn.\\Tech.\\\cite{Autogluon}\end{tabular}
                % & \myRot{\method (Ours)} \\ \hline
                %& \begin{tabular}[c]{@{}c@{}}Prior\\hive\\mon.\\e.g.~\cite{padraig, Murphy2016bWSNSB}\end{tabular}
                %& \begin{tabular}[c]{@{}c@{}}Trad.\\time\\series\\e.g. ~\cite{Akoglu2010EVENTDI,  Castagli92Nonlinear}\end{tabular}
                %& \begin{tabular}[c]{@{}c@{}}Deep\\Learn-\\ing\end{tabular}
                %& \begin{tabular}[c]{@{}c@{}}Symb.\\Reg.\end{tabular}
                & \begin{tabular}[c]{@{}c@{}}\myRot{\method (Ours)}\end{tabular}\\ \hline
            \domainSpec  & \myck  &        &          & \myck \\
            \effective    &        & ?      & ?       &  \myck \\
            \explainable & ?      &   &   &  \myck \\
            scalable    &        & \myck       & ?      &  \myck \\
            %\interpretable & ? & ? & & \myck \\
        \end{tabular}%\vspace{5pt}
        \caption{\label{tab:salesman} \myTag{\method wins},
        fulfilling all the specifications. `?' means `maybe, depending on implementation'. } \vspace{-20pt}
    \end{center}
\end{table}

\smallskip\noindent\textbf{Traditional time series forecasting.}
Time series have attracted huge research interest~\cite{Akoglu, Lexiang, Yasuko}.
%for finance ~\cite{Jacob1999FinTimeAF, papdimitriou2006}; climate ~\cite{Romani2010}; medicine ~\cite{Fischer1999, Lowe1999TreatmentOB};
%for anomaly detection ~\cite{Akoglu2010EVENTDI, He}; 
%web analysis ~\cite{Keogh}; 
%pattern and similarity mining ~\cite{papdimitriou2006, Rafiei}; prediction and forecasting ~\cite{Castagli92Nonlinear}.
%Several books address the topic
%including multiple books
%in the statistics literature
%with multiple textbooks
%~\cite{BoxJ90,Brockwell,Chatfield84Analysis}.
%and numerous papers in the database and data mining literature on indexing, similarity search, and clustering/mining ~\cite{Suzana, Lin2003, Rakthanmanon2012MDLbasedTS, Ye2009} including the earlier works of the authors ~\cite{Christos1994, Matsubara, Matsubara2015, Matsubara2014, Papadimitriou2003, Papadimitriou2005, Yi2000OnlineDM}. 
Classical methods, such as (i) autoregression (AR) based methods with exogenous variable~\cite{Box15}, e.g., ARX/ARMAX/ARIMAX and seasonal ARX/ARMAX/ARIMAX, (ii) exponential smoothing models (ETS)~\cite{Hamilton}, e.g., Holt-Winters, can be used for forecasting but they are neither domain-informative nor explainable. The hyperparameters of these models do not give any insights into the hive's current condition.

\smallskip\noindent\textbf{Deep learning techniques.} 
Such recent methods include
% Recently, Amazon AWS has launched 
Autogluon-Timeseries~\cite{Autogluon} (an open-source autoML library for time-series forecasting),
% an open-source autoML library for easy and accurate deployment of popular DL frameworks for time-series forecasting such as 
DeepAR~\cite{Salinas}, Temporal Fusion Transformer~\cite{Bryan}, etc. 
The main drawback of these methods is the lack of explainability,
despite recent attempts~\cite{Saeed}.
%The drawback of all deep-learning techniques is not being explainable. Though there has been development in the field of explainable AI~\cite{Saeed}, it is not mature enough to be used in real-life scenarios.

\smallskip\noindent\textbf{Contrast to existing literature.} Existing hive health monitoring systems are supervised in nature based on either rule-based~\cite{Murphy2016bWSNSB} 
or black-box methods~\cite{Zhange2015}. %(e.g., simple autoencoders etc.). 
To our knowledge, there exists no prior work to address this problem in an unsupervised manner. Also, none of them except \myTag{\method}  fulfill all specifications
of Table~\ref{tab:salesman}.

\subsection{DataSet and Experimental Setup} \label{sec:DataSet}

%We have collected two datasets from different hive settings (control and treated) and timelines and seasons (Late Summer 2021 and Late Spring 2023).

%\smallskip\noindent\textbf{2021 Data:} 
We have used a data set collected from August to September 2021 (late Summer) from an apiary in UC Riverside, CA, USA.
%\footnote{The data set is available upon request. Please contact the authors or the CIBER bee research center at UC Riverside (\url{https://ciber.ucr.edu/}).} 
This period in that region has very hot and dry climate conditions, with peak temperature exceeding $40^\circ$C. Such conditions can cause significant heat stress %even if the hive colony does not collapse. 
but doesn't lead to colony collapse as long as bees are capable of managing in-hive temperature.

A total of ten homogeneous Langstroth hives with single-box brood chambers were used. They were divided equally into two groups for experimental purposes: {\em control hives} and {\em treated hives}. For each group, hives were selected randomly. Five control hives received no treatment throughout the experiment period. The remaining five hives were treated to combat extreme temperature by placing approximately six kilograms of ice on top of them when the environment temperature exceeded $35^\circ$C (Details in Section~\ref{sec:eval_segmentation}, Figure~\ref{fig:segmentation}).

For data collection, we used a commercial multi-sensor device, Nordic Thingy:52 \cite{thingy52}, which has wireless communication to transfer collected data. %Finally, we inserted it in the middle frame of the hives where the queens and larvae were present. We took out the box with the sensor in regular intervals (Table \ref{tab:daily_log}) to recharge it and replaced it again the next day. 
We recorded temperature %, humidity, air pressure, volatile component, and eCO\textsubscript{2} 
per 20-minute interval for two months (August - September 2021). Every two weeks throughout that period, hives were opened twice: (i) for health evaluation and taking out the sensor device for battery recharging and (ii) for reinstalling the recharged device the next day. While all hives maintained the core temperature within $33 - 36^\circ$C, the control ones had a much higher deviation over several days, implying affected by heat stress. %suggesting comparatively more stress 
(Figure  \ref{fig:curveFitting}). %We placed \textit{Thingy} inside ten hives. Among them, five were control hives that received no treatment throughout the whole experiment period. The rest five were treated to combat the extreme temperature. Six quarts of ice were placed on top of those hives when the environment temperature exceeded 35\textsuperscript{0}C. 
%We use this dataset for all experiments in Section \ref{sec:eval}.

%\smallskip\noindent\textbf{2023 Data:} A data set collected from a control hive from July 10 to August 3, 2023, from the same apiary and hive setting. Due to abundant resources, colonies often produce new queen bees in this period. In case of unfavorable conditions, bees may abscond the hive to start a new colony and avoid a collapse.
%this period is suitable for the emergence of new queen bees. Among many factors, beekeepers may notice bees swarming or absconding from the hive depending on the new queen's reproduction ability.
%We recorded temperature %, humidity, and dew point 
%per 30-minute interval for two months. Hive inspection was done twice: (i) at the beginning of the experiment and (ii) three weeks following the first inspection.  This dataset is only used to demonstrate the effectiveness of our proposed segmentation algorithm, \myAlgorithm.

\smallskip\noindent\textbf{Sensor Placement.} We kept the device inside a box so that bees do not build propolis around the hardware and affect its sensing ability. Each hive had two sensor devices placed in the following locations (Figure \ref{fig:work_flow}(a)): 
\begin{tight_enumerate}

    \item Peripheral Area: On top of the outer frame near the food and honey preservation area.
    \item Core Area: On the central frame near the brood area, where developing broods and the queen are present. The temperature of this area is crucial to hive health as developing broods are susceptible to heat stress, and healthy hives maintain the core area temperature between $33 - 36^\circ$C.
\end{tight_enumerate}

%For our \myTag{\method} method, we use the temperature data from the peripheral and core areas: $\thetaPeri$ and $\thetaCore$, respectively. 
We also installed the sensor device in an empty Langstroth box. As that hive had no bees inside to regulate the core temperature, we consider the reference temperature to be the outer surface temperature of the hives, $\thetaExt$. %and we assume it to be the same for all surfaces regardless of their type. The peripheral temperature, $\thetaPeri$, of treated hives is the surface temperature with ice quarts during treatment days. We consider $\thetaExt$ to be the temperature of the rest of the hive surfaces. 

%\begin{figure}
%	\centering
%	\subfloat[Control Hive]{
%		\includegraphics[clip,width=1\linewidth]{Figures/control_sample.pdf}
%	}\\\vspace{5pt}
%	\subfloat[Treated Hive]{
%		\includegraphics[clip,width=1\linewidth]{Figures/treat_sample.pdf}
%	}
%	\caption{Samples of collected sensor data (temperature). Blue, red and yellow line respectively corresponds to the core temperature, peripheral temperature, and external temperature of the hives.}
%	\label{fig:hive_sensor}	
%\end{figure}

%A sample of our collected data from each type of hive is shown in Figure \ref{fig:hive_sensor}.
 %As shown in this figure, the treated hive maintained the core temperature to stay within the range of ideal temperature ($35 \pm 0.5^\circ$C) whereas the control one had a serious deviation over several days, suggesting stress in the hive.
%We can observe that variations in temperature are significantly larger outside compared to inside the hive, but both core and external temperatures are correlated. In addition,

\subsection{Proposed Method: \method}\label{sec:model}

Table ~\ref{tab:symbols} summarizes the symbols and their definitions used in the paper.
%\vspace{-12pt}
\begin{table} [ht]
    \begin{center}
    \small
        \begin{tabular}{p{2cm} | p{5.5cm}}
            Symbols & Definitions\\
            \hline \hline
            $\mathcal{D}$ & Temperature dataset \\
            $\mathcal{M}$ & Model in Eqn.\ ~\eqref{eq_1}\\
            $\mathcal{L(D|M)}$ & Likelihood of $\mathcal{D}$ given $\mathcal{M}$ \\
            $\thetaIdeal$ & Ideal core temperature \\
            
            $\thetaExt$  & External temperature \\

            $\thetaPeri, \thetaCore$  & Hive's Peripheral \& Core temp. \\
            
            $\thetaExtNorm$ &  $\thetaExt$ relative to $
            \thetaIdeal$ \\

            $\thetaPeriNorm, \thetaCoreNorm$ &  $ \thetaPeri, \thetaCore$ relative to $
            \thetaIdeal$ \\
            
            $\hat{\theta}(t)$ & Fitted/forecasted $\thetaCoreNorm$ relative to $\thetaIdeal$ \\
            
            $\thetaAdj$ & Second surface temp.\ based on hive type \\
            
            $\strengthCooling$, $\strengthHeating$ & Bees' strength of cooling \& heating\\
            
        \end{tabular} %\vspace{-5pt}
        \caption{\label{tab:symbols} Symbols \& Definitions }
    \end{center}
\end{table}
\vspace{-15pt}
\subsubsection{Proposed Model: \methodTwo.}\label{sec:proposed_model}
We present our principled model, \methodTwo, to monitor and forecast the hive core temperature $\thetaCore$, given the outside temperature $\thetaExt$.

%\begin{figure}
%	\centering
%	\subfloat[]{
%		\includegraphics[height =2.5 cm, width = 4cm]{Figures/periphery_sensor.pdf}
%	}
%	\subfloat[]{
%		\includegraphics[height =2.5 cm, width = 4cm]{Figures/core_sensor.pdf}
%	}\vspace{-10pt}	
%	\caption{Installing sensors onto the hive: (a) On the outer frame (Periphery area); (b) On the center frame (Core area).}\vspace{-10pt}	
%	\label{fig:hive_sensor_placement}	
%\end{figure}

\begin{lemma}%[\methodTwo for Control Hive]
Given the external temperature, $\thetaExt$, and the strengths $\strengthHeating$ and $\strengthCooling$ of honeybees for warming up and cooling down the hive, respectively, the hive core temperature, $\thetaCore$, would obey:
%\vspace{-3pt} 
\begin{equation}
  \resizebox{\hsize}{!}{\boxed {\frac{\partial {\thetaCoreNorm} }{\partial t} =\begin{cases}
    \thetaExtNorm + \thetaAdjNorm - 2\thetaCoreNorm - \strengthCooling \thetaCoreNorm, &
    \text{if $ \thetaExtNorm \geq 0$.}\\
    \thetaExtNorm +\thetaAdjNorm - 2\thetaCoreNorm + \strengthHeating \thetaCoreNorm, & \text{otherwise.}
  \end{cases}} }
\label{eq_1} 
\end{equation} 
\end{lemma}\vspace{-8pt} 
where 
\vspace{-10pt} 
\begin{equation*}
  \thetaAdjNorm = \begin{cases}
    \thetaExtNorm, &
    \text{if $hive \ type = control$}\\
    \thetaPeriNorm, &
    \text{if $hive \ type = treated$}.
  \end{cases}
\end{equation*} 
% \vspace{-6pt} 
Next, we provide the justification of our model, \methodTwo (Eqn.~\ref{eq_1}) by using 
first principles: Thermal diffusion and Control theory.

\subsubsection{Justification}
% \begin{proof} %\renewcommand{\qedsymbol}{$\blacksquare$}
Here, all temperatures are relative to the hive's ideal core temperature, 
$\thetaIdeal$, as bees efforts to keep the core temperature, $\thetaCore$, within the normal range, depends on its deviation from $\thetaIdeal$. It is worth noting that the ideal temperature $\thetaIdeal$ is within $33 - 36^\circ$C but the exact value can vary slightly within this range, from day to day and from hive to hive, due to several reasons, such as hive size, population, and season.
%fatigue from previous days.

To take into account the effect of treatment, we consider two types of hive surface temperatures: top ($\thetaAdj$) and the rest surfaces ($\thetaExt$). For control hives, we assume all surfaces share the same $\thetaExt$ since all of them directly face the outside. For treated hives, there was ice on the top surface on hot days, and the top surface temperature, $\thetaAdj$, would be different from the rest surfaces and can be obtained by the sensor installed in the peripheral area, $\thetaPeri$. Based on this, we derive Eqn.\ \eqref{eq_1} in the following steps.
%In Eqn.\ \eqref{eq_1}, $\thetaAdj$ is used to model the second surface depending on hive type: the top (treated hives with ice) and the rest. Since we placed only one sensor on the outer frame of the hive, we assume the same $\thetaExt$ for all surfaces of control hives. As there was ice on top of treated hives on warmer days, the top surface temperature, $\thetaPeri$, would be different from the rest of the surfaces. So, we modify Eqn.\ \eqref{eq_1} accordingly for different types of hives.
%The first term in Eqn. \eqref{eq_1} is due to heat transfer, and the second term is the feedback loop: when the core temperature of the hive, $\thetaCoreNorm$ is away from $\thetaIdeal$, the bees try to bring it to zero with strengths $\strengthCooling$ and $\strengthHeating$. Now let's take an elaborate look at the equation.

\smallskip\noindent\textbf{Step 1: Physics - Thermal Diffusion.}
The first term in Eqn. \eqref{eq_1} is due to heat transfer: for an empty hive, the rate of change in hive core temperature,
$\frac{\partial {\thetaCoreNorm}}{\partial t}$ obeys the thermal diffusion model \cite{Crank75}, 
%\vspace{-6pt}
$$ \boxed{\nabla ^ 2 {\theta}  = \frac{\partial^2 {\theta} }{\partial x^2} + \frac{\partial^2 {\theta} }{\partial y^2} + \frac{\partial^2 {\theta} }{\partial z^2} \propto \frac{\partial {\theta} }{\partial t}}$$

So, we can define $\frac{\partial {\thetaCoreNorm} }{\partial t} = \thetaExtNorm + \thetaAdjNorm - 2\thetaCoreNorm$. 

\smallskip\noindent\textbf{Step 2: Control Theory - (split) P-controller.}
The second term is the feedback loop \cite{Nise19}: when the hive core temperature, 
$\thetaCore$ is away from ideal core temperature $\thetaIdeal$, bees try to bring it to zero with strengths $\strengthCooling$ and $\strengthHeating$ and their efforts are proportional to $\thetaCoreNorm$ i.e., if bees are left alone, 
%\vspace{-6pt}
$$
    \frac{\partial {\thetaCoreNorm} }{\partial t} = \begin{cases}
    -\strengthCooling \thetaCoreNorm, &
    \mathrm{if~} \thetaExtNorm \geq 0.\\
    + \strengthHeating \thetaCoreNorm, & \mathrm{otherwise}.
  \end{cases}
$$

Notice that we propose to have a {\em split} in the standard proportional (P) controllers:
the bees have different ability to cool ($\strengthCooling$: when $\thetaExt$ goes above $\thetaIdeal$)
and different ability to heat ($\strengthHeating$: when $\thetaExt$ goes below $\thetaIdeal$) - this is exactly what we call
{\em split} P-controller.
%When $\thetaExt$ goes above $\thetaIdeal$, bees try to maintain $\thetaCore$ by cooling using their strength, $\strengthCooling$. Similarly, when $\thetaExt$ goes below $\thetaIdeal$, bees go into the heating mode using their strength $\strengthHeating$. 
We have followed the sign convention: negative work for heating and positive work for cooling. Combining these two effects makes the equation the same as Eqn. \eqref{eq_1}. %$\quad\blacksquare$
% \end{proof}

%\begin{model}[\methodTwo for Treated Hive]
%Given the external temperature, $\thetaExt$, and the strengths $\strengthHeating$ and $\strengthCooling$ of honeybees to react for warming up and cooling down the hive, respectively, the core temperature of a treated hive, $\thetaCore$, would obey:
%\begin{equation}
%  \boxed {\frac{\partial {\thetaCoreNorm} }{\partial t} =\begin{cases}
%    \thetaExtNorm + \thetaPeriNorm - 2\thetaCoreNorm - \strengthCooling \thetaCoreNorm, \\
%    \hspace{125pt} \text{if $ \thetaExtNorm \geq 0$}.\\
%   \thetaExtNorm + \thetaPeriNorm - 2\thetaCoreNorm + \strengthHeating \thetaCoreNorm, \\ \hspace{130pt}\text{otherwise}.
%  \end{cases}} \label{eq_2}
%\end{equation}
%\end{model}

%Here, $\thetaPeriNorm$ is the peripheral temperature relative to $\thetaIdeal$. 
Our goal is to find out such cooling and heating strengths $\strengthCooling$, $\strengthHeating$, and $\thetaIdeal$ that best reconstructs $\thetaCore$, given $\thetaExt$ and $\thetaAdj$.    

%\subsection{Segmentation and Event Detection}

\subsubsection{Proposed Algorithm: \myAlgorithm.} \label{sec: algorithm}
Based on \methodTwo, we develop a segmentation algorithm to find out a set of segments and their parameters by penalizing model complexity. The cut-points spot discontinuity corresponding to events affecting bees' thermoregulation ability. We use AIC (Akaike Information Criterion) to measure the goodness of fit of our algorithm. We want to minimize the following equation:
%\vspace{-3pt}
\begin{equation}
    AIC = -2ln \mathcal{L}(\mathcal{D}|\mathcal{M}) + 2(m + (m+1)p) \label{eq_6}
\end{equation}
%\sum_{i = 1}^N ln f(\theta_{core}[i] - \hat{\theta}_{core}[i])
where $ \mathcal{L}(\mathcal{D}|\mathcal{M})$ is the likelihood of data $\mathcal{D}$ given the model $\mathcal{M}$ described in Eqn.\ \eqref{eq_7} and $m, p \,(=3)$ represent the number of cut points and the number of parameters for each segment respectively. 
%\vspace{-6pt}
$ \mathcal{L}(\mathcal{D}|\mathcal{M})$ is given by:
\begin{equation}
    \mathcal{L}(\mathcal{D}|\mathcal{M}) = \prod_{i = 1}^N f(\Theta[i] - \hat{\Theta}[i]) \label{eq_7}
\end{equation} 
where $f(.)$ is the pdf of a continuous normal random variable, and $N$ is the number of total time ticks.

We want to find the combination of segments and corresponding parameters, $\{\strengthCooling, \strengthHeating, \thetaIdeal\}$ that minimizes Eqn.\ ~\eqref{eq_6} and outputs the best-fitted core temperature, $\thetaCoreFit$.
Algorithm \ref{algo:1} demonstrates \myAlgorithm, the implementation of \methodTwo for curve fitting and segmentation. 
%using the recorded daily temperature sequences $\thetaExt$, $\thetaPeri$ and $\thetaCore$. 
 %As the simplified model (\methodOne given in Sec.~\ref{sec:simplified_model}) is similar to our full model (\methodTwo in Sec.~\ref{sec:full_model}), we are only showing the case for \methodTwo.
%\vspace{-18pt}
\begin{algorithm2e}
\DontPrintSemicolon
\caption{\myAlgorithm \label{algo:1}}
\KwIn{$\mathcal{D} = \{\mathcal{\AllthetaExt}, \mathcal{\AllthetaPeri}, \mathcal{\AllthetaCore}\}$} 
\KwOut{\{$\mathcal{\CutPoint}$, $\mathbfcal{X}$\}}
%\KwOut{\begin{minipage}[t]{0.8\textwidth} %  choose a suitable width
%\begin{enumerate}[nosep,left=0pt,label=(\alph*)]
%\item Position of cut-points $\mathcal{\CutPoint} = \{cp_1, cp_2, ...,cp_m\}$
%\item Set of parameters for all segments 

%$\mathbfcal{X} = \{\mathcal{X}_1, \mathcal{X}_2, ... \mathcal{X}_{m+1}\}$
%\end{enumerate}
%\end{minipage}}

/* Fitting: compute $\thetaCoreFit$ */\label{ln:fitting_begin}

\lFor{each pair $(e,d) \in (\mathcal{P},\mathcal{D})$}{\label{ln:search_space}
    sol. Eqn.\ \eqref{eq_1}
}\label{ln:fitting_end}

/*Segmentation and finding cut points*/\label{ln:seg_begin}

$j \gets 1, \mathcal{\CutPoint} \gets \{0, n\}, \mathbfcal{X} \gets \emptyset$ 

\While{j $<$ n %\text{and} $\mathcal{L}_{min} > \mathcal{L}_{min, prev}$
}{
    \For{each point, $p \in \{\mathcal{D} - \mathcal{\CutPoint}\}
    %\notin \mathcal{\CutPoint}
    $}{
        $cp \gets \{\mathcal{\CutPoint}\} \cup p$  //set of cut-points
        
        \For{i = 0 to length($\mathcal{\CutPoint}$)}{
            Create Segment $s[i] \gets \mathcal{D}[cp[i]:cp[i+1]]$

           %Do fitting, $\theta_{fitted} \gets$ ($\mathcal{P}_{RMSE_{min}}, s[i]$)

           Fit and choose $\mathcal{P}_{best} \gets \mathcal{P}_{RMSE_{min}}$
        }

        Calculate $\mathcal{L}(\thetaCore, \hat{\Theta}(t))$ /*Eqn. \eqref{eq_7}*/ 
    }

    Choose $\mathcal{L}_{min}(\thetaCore, \hat{\Theta}(t))$, and $\mathcal{P}_{\mathcal{L}_{min}}$
    
    %\lIf{$\mathcal{L}_{min} > \mathcal{L}_{min, prev}$}{
    %    break
    %}

    {$\{\mathcal{\CutPoint}, \mathbfcal{X}\} \gets \{ 
    \{\mathcal{\CutPoint}\} \cup p_{\mathcal{L}_{min}}, \{\mathbfcal{X}\} \cup \mathcal{P}_{\mathcal{L}_{min}}\}$}

    %$\mathbfcal{X} \gets \{\mathbfcal{X}\} \cup \mathcal{P}_{\mathcal{L}_{min}}$ 
    
    $j \gets j + 1$
}\label{ln:seg_end}

%\Return  $\{\mathcal{\CutPoint}, \mathbfcal{X}\}$ 
\end{algorithm2e} 

\myAlgorithm takes $\mathcal{D} = \{\mathcal{\AllthetaExt}, \mathcal{\AllthetaPeri}, \mathcal{\AllthetaCore}\}$ as input and $\mathcal{CP} = \{cp_1, cp_2, ... cp_m\}$, $\mathbfcal{X} = \{\mathcal{X}_1, \mathcal{X}_2, ... \mathcal{X}_{m+1}\}$ as output:

\begin{tight_itemize}
    \item $\mathcal{D} = \{\mathcal{\AllthetaExt}, \mathcal{\AllthetaPeri}, \mathcal{\AllthetaCore}\}$ is the dataset for a hive. Here, $\mathcal{\theta_E}$, $\mathcal{\theta_P}$, $\mathcal{\theta}$ are sets of daily external, peripheral 
    %Here, $\mathcal{\theta_E} = \{\theta_{ext,1}(t), \theta_{ext,2}(t) .. \theta_{ext,n}(t)\}$ is the set of daily external temperature sequences over the experiment period where $\theta_{ext,i}(t)$ is the external temperature for i-th day, i.e., $1 \leq i \leq n$, $n$ is the total count of sequences for the hive. Similarly,  $\mathcal{\theta_P}$ and $\mathcal{\theta_C}$ are the sets of daily peripheral %temperature sequences 
    and core temperature sequences over the experiment period.
    \item $\mathcal{CP} = \{cp_1, cp_2, ... cp_m\}$ and $\mathbfcal{X} = \{\mathcal{X}_1, \mathcal{X}_2, ... \mathcal{X}_{m+1}\}$ are the sets of cut point locations and $\{\strengthCooling, \strengthHeating, \thetaIdeal\}$ respectively, where $m$ is the number of cut points i.e., $m \leq n$, $n$ is the number of daily sequences. 
    %and $\mathcal{X}_i = \{\strengthCooling, \strengthHeating, \thetaIdeal\}_i$ is the parameters for i-th segment i.e., $1 \leq i \leq (m+1)$.
\end{tight_itemize}

\smallskip\noindent\textbf{Fitting.} (lines~\ref{ln:fitting_begin} to \ref{ln:fitting_end}) We use Levenberg-Marquardt (LM) optimization for faster convergence (other kinds of optimization or linear search can also be used) %over different values of $\thetaIdeal$ and $(\strengthCooling, \strengthHeating)$ for \methodTwo. For defining the search space, we considered the following constraints:
considering the following constraints for the search space $\mathcal{P} = \{\mathcal{S_C}, \mathcal{S_H}, \mathcal{\theta_I}\}$ (line~\ref{ln:search_space}).

\begin{tight_enumerate}
    \item $\mathcal{S_C} \land  \mathcal{S_H} = [0, s_\infty]$
    are the search space for bees' cooling and heating strength, 
    $\strengthCooling$ and $\strengthHeating$ respectively. %where $\strengthInfinity$ is the upper limit for the user-defined strength of bees.
    Normally, $\mathcal{S_C}$ and $\mathcal{S_H}$ vary between [0,50] (Details in Section~\ref{sec:explainability}) with a few exceptions. So, the upper limit $s_\infty$ can be any value above 50.

    \item $\mathcal{\theta_{I}}= [\theta_{ideal}^{lower}, \theta_{ideal}^{upper}]$ is the search space for $\thetaIdeal$ that 
    can be anywhere between $33-36^\circ$C for healthy hives and go outside this range for unhealthy ones. %As the experimental hives were healthy %(except in a few cases, according to the domain specialist) 
    We assumed the range to be $(33-36)\pm 2^\circ$C. 
    
\end{tight_enumerate}
%Considering these constraints, we define $\mathcal{P} = \{\mathcal{S_C}, \mathcal{S_H}, \mathcal{\theta}\}$ (line~\ref{ln:search_space}), the entire search space of parameters for our algorithm. Here, $\mathcal{S_C} = \{\strengthCooling_1, \strengthCooling_2, ... \strengthCooling_a\}$ and $\mathcal{S_H} = \{\strengthHeating_1, \strengthHeating_2, ... \strengthHeating_a\}$ are the search space for the strength of bees for cooling and heating, respectively, where $\strengthCooling_i$ and $\strengthHeating_i$ are the i-th instances of the search spaces,  i.e., $0 \leq \strengthCooling_i \land \strengthHeating_i \leq \strengthInfinity$, where $\strengthInfinity$ is the upper limit for the strength of bees that can be defined by the user.

%The last parameter $\mathcal{\theta} = \{\theta_1, \theta_2, ... \theta_b\}$ is the search space for $\thetaIdeal$, where $\theta_i$ ($1 \le i \le b$) is the i-th instance of the area, i.e., $\theta_{ideal}^{lower} \leq \theta_{i} \leq \theta_{ideal}^{upper}$. Here, $\theta_{ideal}^{lower}$ and $\theta_{ideal}^{upper}$ are the lower and upper bounds and can be chosen like $s_\infty$.

There can be two instances when we cannot calculate the strength of bees and do either linear interpolation or forward/backward filling, considering strength(s) as a continuous variable(s).:

\begin{tight_itemize}
    \item Case 1: $\thetaCore$ is almost constant the whole day ($\strengthCooling$, $\strengthHeating$ will reach the upper limit).

    \item Case 2: $\thetaCore$ is below/above $\thetaIdeal$  the whole day($\strengthCooling$/$\strengthHeating$ will be zero or any other constant).
\end{tight_itemize}

%In such cases, we can do linear interpolation %(strengths changes linearly over time)  forward/backward filling 
%(strengths for the current sequence remain the same as the previous sequence/day i.e., strength is a continuous variable).

%Though we used linear search for all our experiments, any optimization (e.g., Newton-Rapshon, Levenberg Marquardt (LM) optimization) can be used to reduce the computation time required to fit the curve and find out the cut points.

\smallskip\noindent\textbf{Segmentation and finding cut-points.} (lines~\ref{ln:seg_begin} to ~\ref{ln:seg_end}) 
For this part, we use the following notations.
\begin{tight_itemize}
\setlength\itemsep{0em}
   \item $\mathcal{L} = \{l_1, l_2, ... l_{n}\}$ is the set of likelihoods for different cut-points positions and $\mathcal{L}_{min}$ is the minimum of $\mathcal{L}$.
    %\item $\mathcal{L}_{min,prev}$ is the minimum of $\mathcal{L}_{prev}$, likelihood calculated for previous iteration.
    \item $\mathcal{P}_{RMSE_{min}}$ and $\mathcal{P}_{\mathcal{L}_{min}}$ is the parameters $\mathcal{X} = \{\strengthCooling, \strengthHeating, \thetaIdeal\}$ 
   for each segment that gives minimum RMSE and AIC for $\mathcal{D}$.
\end{tight_itemize}
For segmentation, we follow a greedy approach. Let's begin with the simplest case: a single cut-point search with two segments (i.e., \ $\mathcal{S}_1$ and $\mathcal{S}_2$) for a given dataset $\mathcal{D}$. Each segment will have its respective parameters, $\mathcal{X}_i = \{\strengthCooling, \strengthHeating, \thetaIdeal\}_i$ for best reconstruction. We start with the cut-point position, $cp = 1$ and iterate till $n$. %, the total count of sequences. 
For each cut-point, we will have $\{(\mathcal{S}_1,\mathcal{X}_1), (\mathcal{S}_2, \mathcal{X}_2)\}$ and $\mathcal{L}$ for $\mathcal{D}$. We will choose $\mathcal{\CutPoint} = \{cp_1\}$ with  $\mathcal{L}_{min}$. How about a second one? We will repeat the above process, excluding $cp_1$ from our search. Then, we will have $\mathcal{\CutPoint} = \{cp_1, cp_2\}$ and repeat the same process for the next one till we find reconstruction with minimum AIC. %For the third one, we will exclude $cp_1$ and $cp_2$ and repeat the whole process. For each cut-point search, we will store the value of AIC and compare it with the previous step (i.e., AIC with two cut-points and a single cut-point). We will stop searching for the next cut-point when we find the minimum AIC. 

Reconstruction using more segments will be closer to the original reconstruction. As will be shown in  Figure \ref{fig:curveFitting},
we get the best fit if  
each sequence is considered independent of the other i.e., $n$ sets of $\mathcal{X}$ for n sequences. But more segments also increase the chances of detecting cuts with no significant changes in $\mathcal{X}$.
In segmentation, we aim to reduce the model complexity by reducing the number of $\mathcal{X}$s. Thus, instead of considering minimum AIC, we use the combination of segments and parameters that gives 10\% higher AIC than the minimum one \cite{Yasuko}. The experimental results in Section~\ref{sec:eval_segmentation} demonstrate that our assumption is reasonable.

\section{Empirical Evaluation}\label{sec:eval}
This section presents the effectiveness of %both \methodOne and 
our proposed method, \method, using our collected datasets and experiments designed to ensure the following:

%We design our experiments to ensure the following:

\begin{enumerate}[start=1,label={Q\arabic*}]
\setlength{\itemsep}{0pt}
\setlength{\parskip}{0pt}
\setlength{\parsep}{0pt}
\item \textit{\effectiveness:} How accurately does our proposed method \method detect events affecting hive health and forecast the core temperature, $\thetaCore$? (Sec.\ \ref{sec:eval_segmentation} \& \ref{sec: forecasting})
\item \textit{\explainability:} %Can our segmentation algorithm, \myAlgorithm explain detected cut points well?
%Can our method explain (i) the change in bee strengths in detected segments and (ii) the effect of ice treatment on bees' thermoregulation ability well? (Sec. \ref{sec:explainability})
Can \method explain the change in bee strengths and the effect of corresponding events in detected segments? (Sec.\ \ref{sec:explainability})
\item \textit{\scalability:} Does our model perform linearly in time? (Sec. \ref{sec:scalability})
\item \textit{\observations:} Do the observations based on our experiments coincide with the domain experts' expectations? (Sec. \ref{sec:observation})
\end{enumerate}

All experiments are done on a Dell XPS 15 laptop with an Intel i9 processor and 32GB memory. We used averaged hourly data for all experiments as the change in temperature is noticeable if considered hourly. There are missing data points due to the nature of real-world data collection, e.g., sensors not working properly, beekeepers' intervention for inspection, etc. However, no interpolation is done since $\myAlgorithm$ is robust to handling missing values. Due to space constraints, we present one example hive from each hive setting in this section, but our findings are applicable to all hives used during the experiment.

\subsection{Q1 - Effectiveness: Event-detection.}\label{sec:eval_segmentation}
\begin{figure*}
    %\subfloat[Control Hive (cut and fit using \methodOne)]{
    %    \includegraphics[height = 3cm, width = 8.5cm]{Figures/model_1_265_seg.pdf}
    %}
    \subfloat[Control Hive (High Heat Stress)]{
        \includegraphics[height = 3.75cm]{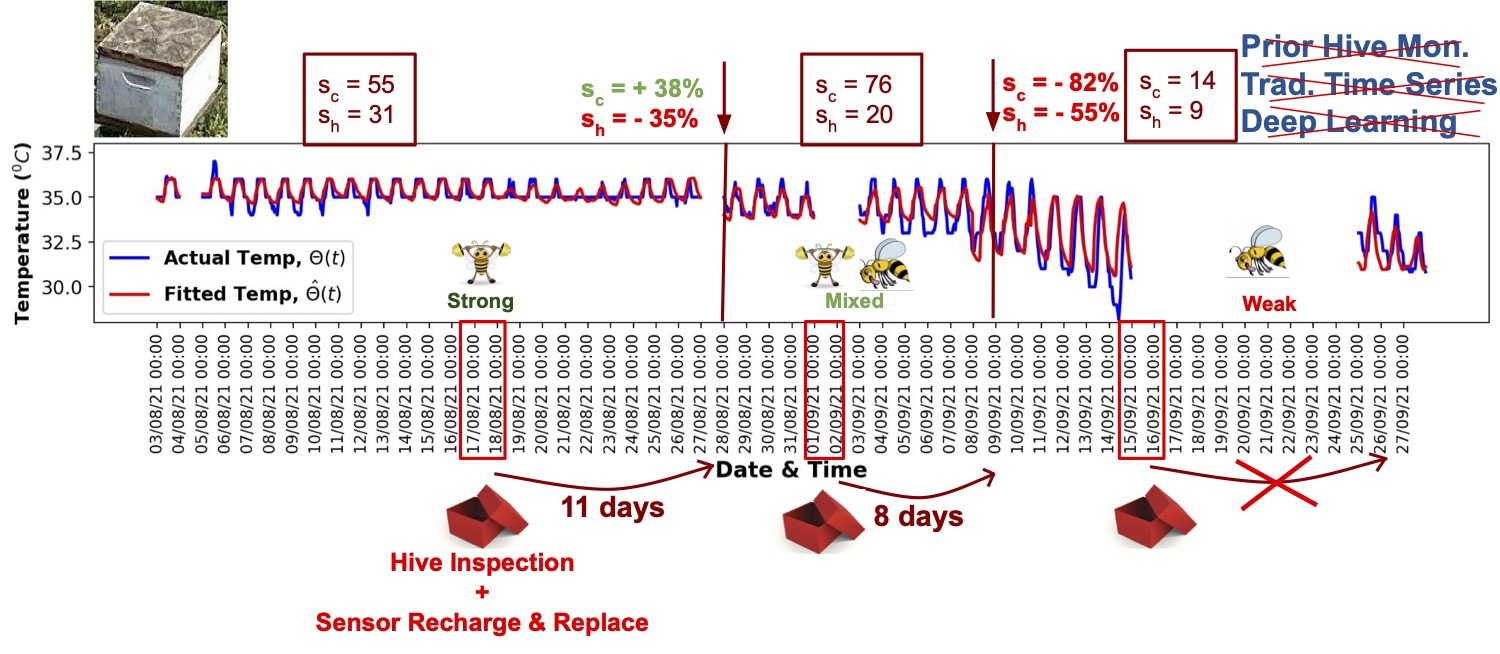}
    }%\\\vspace{10pt}
    %\subfloat[Treated Hive (cut and fit using \methodOne)]{
    %    \includegraphics[height = 3cm, width = 8.5cm]{Figures/model_1_260_seg.pdf}
    %}
    \subfloat[Treated Hive (Low Heat Stress)]{
                \includegraphics[height = 3.75cm]{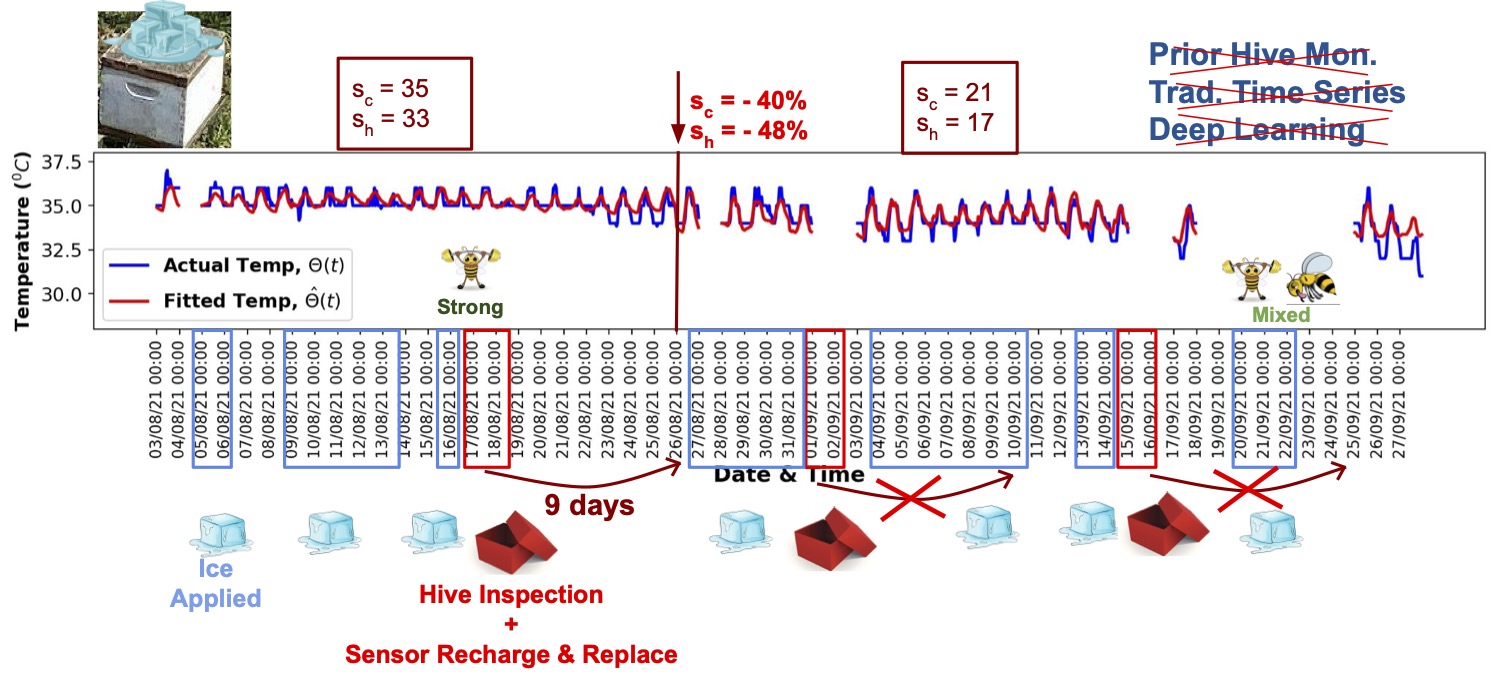}

    }\vspace{-3pt}
    \caption{\myTag{\method is effective \& explainable:} Accurate segments: \myAlgorithm fits (red line) the data (blue line) well and finds cut-points (brown vertical line) followed by unexpected events (red boxes: intensive hive inspection, sensor recharge and replace). 
    Notice that the baselines are not applicable (cannot do segmentation).}
    \label{fig:segmentation}
\end{figure*} 

%\begin{figure}
    %\subfloat[Control Hive (cut and fit using \methodOne)]{
    %    \includegraphics[height = 3cm, width = 8.5cm]{Figures/model_1_265_seg.pdf}
    %}
%    \includegraphics[height = 3.5cm, width = 8 cm]{Figures/SW_15_SR_4_seg.jpg}
%    \caption{\myTag{\method is effective \& explainable:}  Case Study on 2023 Data: \myAlgorithm fits (red line) the data (blue line) well and finds cutpoints (brown vertical line) that explain events reported during inspection (orange boxes: regular inspection). 
    %Notice that the baselines do not apply (can not do segmentation).
%    }
%    \label{fig:segmentation_2}
%\end{figure} 

%To demonstrate the effectiveness of \myTag{\method}, we consider all events reported during hive inspections and show \myAlgorithm is able to detect the change in thermoregulation following the events:

%\smallskip\noindent\textbf{Case Study on 2021 Data:}
\noindent\textbf{Experimental Setup.}
Frames were taken out of hives twice in two days %in a row 
during every inspection to recharge and replace sensors installed inside (Section ~\ref{sec:DataSet}). It takes more time than usual and thus induces more stress and damage to bees and larvae. We refer to this type of inspection as intensive compared to regular ones.
%It is considered intensive compared to regular inspection and can result in injured/dead larvae. 
All hives were opened on the same days: 17-18 August,  1-2 September, and 15-16 September (marked by open red boxes: {\em intensive inspection}). Also, ice treatments during hot days for treated hives are marked by blue boxes with ice symbols (Figure ~\ref{fig:segmentation}). We want to see if \myAlgorithm is able to detect changes in $\thetaCore$ following the inspections.

\begin{tight_enumerate}
    \item Intensive hive inspections: %Bees need to remove injured/dead larvae caused by hive inspection as it may have a negative impact on hive hygiene and population growth. 
    To maintain hive hygiene and population growth, bees need to remove larvae affected during inspection and reestablish homeostasis. For stressed hives, maintaining thermoregulation and cleanliness at the same time can lead to a noticeable increase in fluctuation in $\thetaCore$ within 9-10 days following the inspection.
    \item Effect of ice treatment: Domain experts expect ice treatment during hot days will help maintain stress and thermoregulation following intensive inspections in the long run.
\end{tight_enumerate}

\begin{figure}
    \subfloat[Control Hive]{
        \includegraphics[ height = 3.5cm]{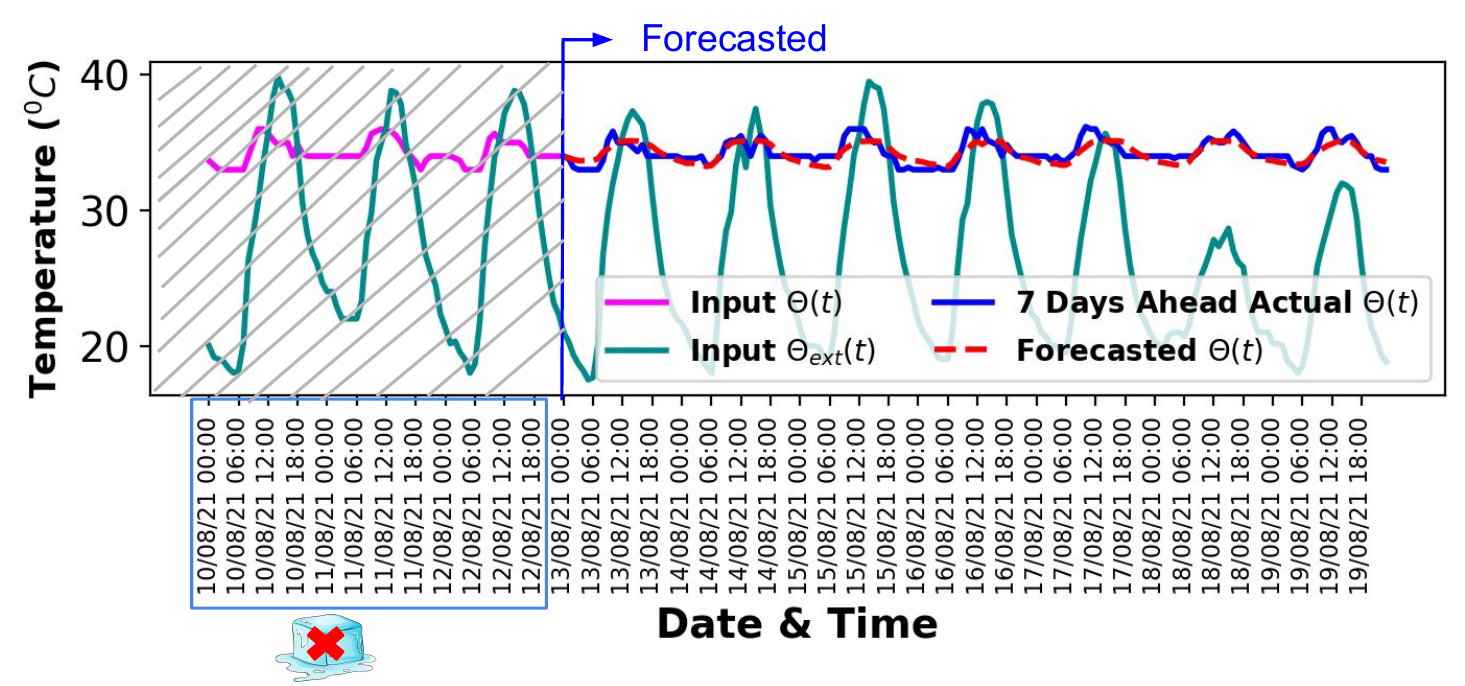}
    } \\
    \subfloat[Treated Hive]{
        \includegraphics[ height = 3.5cm]{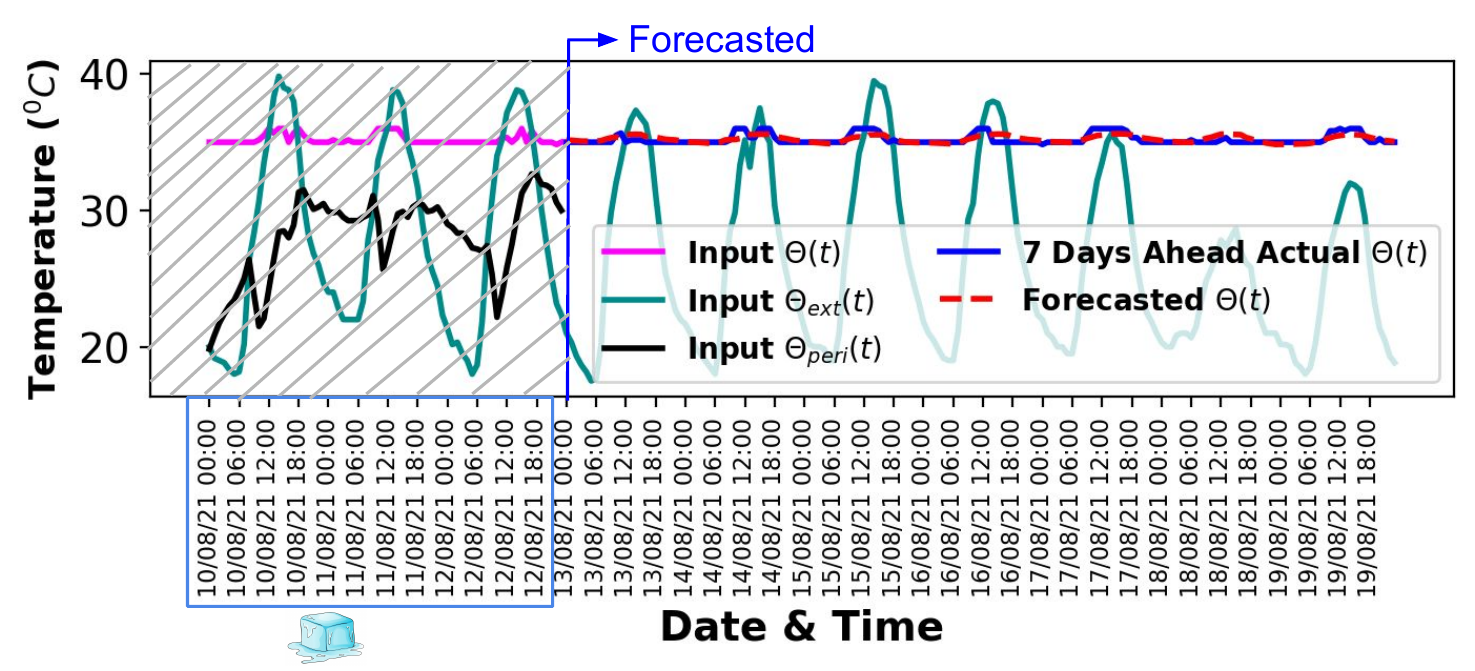}
    }%\vspace{-5pt}    
\caption{\myTag{\method forecasts well}: The seven days ahead forecasting (dotted red line) is close to the actual core temperature (solid blue line).}
\label{fig:forecasting}
\end{figure}

\noindent\textbf{Results.}  \myAlgorithm detected $n =2$ and $n=1$ cut-points and corresponding parameters for each segment (brown vertical line in Figure \ref{fig:segmentation}) for the control and treated hive, respectively. The reconstructed curve (red line) by \myAlgorithm is well aligned with the recorded $\thetaCore$ (blue line). 

For the control hive (Figure \ref{fig:segmentation}(a)), \myAlgorithm detected cut-points on 28 August and 9 September: 11 and 8 days following the first and second inspection, respectively. We suspect it is correlated to hive cleaning (affected larvae removal) and re-establishing homeostasis after intensive inspection. No cut-point was detected after the openings on 15-16 September. The reason might be the recorded data length after the inspection: too short to detect discontinuity. 

For the treated hive, \myAlgorithm detected a cut-point (Figure \ref{fig:segmentation}(b)) on 26 August: 9 days following the first inspection. No cut-points were detected after the next two inspections. This indicates bees in the treated hive were good at managing the aftereffects of intensive inspections. We strongly suspect this is a result of the positive cumulative effect of ice treatment over time. 

In summary, our algorithm \myAlgorithm detected a total of 9 cut-points for 15 inspections (3 per hive) of control hives and 5 cut-points for 15 inspections of treated ones. There were no false positives, and the detected events had ecological significance 
(Section~\ref{sec:explainability}).

\subsection{Q1 - Effectiveness: Forecasting} \label{sec: forecasting} 

% \begin{figure*}
% \begin{minipage}{0.48\textwidth}
%   \centering
%   \includegraphics[height = 4cm, width = 7cm]{Figures/model_1_265_forecasted.pdf}
%   \captionsetup{labelformat=empty}
%   \caption{(A) Control Hive (forecasted using \methodOne)}
% \end{minipage}%
% \addtocounter{figure}{-1}
% \begin{minipage}{0.48\textwidth}
%   \centering
%   \includegraphics[height = 4cm, width = 7cm]{Figures/model_2_265_forecasted.pdf}
%   %need to change this figure (wrong legend)
%   \captionsetup{labelformat=empty}
%   \caption{(B) Control Hive (forecasted using \methodTwo)}
% \end{minipage}
% \addtocounter{figure}{-1}
% \begin{minipage}{0.48\textwidth}
%   \centering
%   \includegraphics[height = 4cm, width = 7cm]{Figures/model_1_260_forecasted.pdf}
%   \captionsetup{labelformat=empty}
%   \caption{(C) Treated Hive (forecasted using \methodOne)}
% \end{minipage}%
% \addtocounter{figure}{-1}
% \begin{minipage}{0.48\textwidth}
%   \centering
%   \includegraphics[height = 4cm, width = 7cm]{Figures/model_2_260_forecasted.pdf}
%   \captionsetup{labelformat=empty}
%   \caption{(D) Treated Hive (forecasted using \methodTwo)}
% \end{minipage}
% \addtocounter{figure}{-1}
% \caption{\myTag{Our models forecast well}: The forecasting (blue red line) of the upcoming seven days is closer to the actual core temperature sequence (solid blue line).}
% \label{fig:forecasting}
% \end{figure*}

\noindent \textbf{Experimental Setup.} %For this experiment, only 2021 Data is used. 
For each trial, we use three days of temperature sequences (endogenous variable: $\thetaCore$ and exogenous variable: $\thetaExt$, $\thetaAdj$) as inputs and forecast $\thetaCore$ hourly for the next seven days, given $\thetaExt$ for the forecasting period. The forecasted core temperature is compared against the actual core temperature using RMSE: RMSE = $ \sqrt{\frac{\sum_{i=1}^{K} (\Theta[i]-\hat{\Theta}[i])^2}{K}}$, where $K$ is the number of forecasted time-ticks and $\hat{\Theta}[i]$ is the forecasted value at $i$-th time-tick.

\smallskip\noindent\textbf{Baselines.} To demonstrate the effectiveness of \method in forecasting, we compared it to the state-of-the-art time series forecasting models ARX, seasonal ARX~\cite{Box15}, Holt-Winters~\cite{Hamilton}, and DeepAR from Autogluon~\cite{Autogluon}. We followed standard practice to determine (i) the order of ARX and seasonal ARX: AIC~\cite{Hurvich}, and (ii) the hyperparameters of the Holt-Winters method: additive trend and seasonality and non-linear optimization. For DeepAR, we used 2 RNN layers with 40 cells for each and a learning rate of 0.001. As Autogluon (DeepAR) requires the input sequence length to be greater than twice the length of the forecasting period, we recursively forecast seven days ahead for three days of input data.

For \method, we use estimated parameters ($\strengthCooling$, $\strengthHeating$ and $\thetaIdeal$) from the input sequences along with $\thetaExt$ for the next seven days to forecast $\thetaCore$ for that time period. 

\smallskip\noindent\textbf{Results.} Figure \ref{fig:forecasting} demonstrates examples of the seven-day-ahead forecasting of $\thetaCore$ (13-19 August 2021) using three days' temperature sequences (10-12 August 2021) as inputs. The magenta, black, and cyan lines represent the input $\thetaCore$, $\thetaExt$, and $\thetaPeri$ (for the treated hive), respectively. % and it aligns well with the blue line.
%First, we calculate $\strengthCooling$, $\strengthHeating$ %($\strength$ for \methodOne), 
%and $\thetaIdeal$ for the input sequence using \methodTwo (See Sec. ~\ref{sec:model}) and use the external temperature $\thetaExt$ %(i.e., environment temperature) 
%of 14-20 August, along with these parameters to forecast $\thetaCore$. As 11-13 August, treatment was applied to treated hives; we consider $\thetaPeri$ as input while forecasting the core temperature of the treated hives. %As we do not have a priori knowledge of $\thetaPeri$ for the forecasting period, only the calculated parameters and $\thetaExt$ are used while forecasting. 
As can be seen, the forecasted core temperature sequence (dashed red line) is very close to the actual one (blue line), demonstrating the effectiveness of our method.

\begin{figure}
\centering
        \includegraphics[height = 4.25cm]{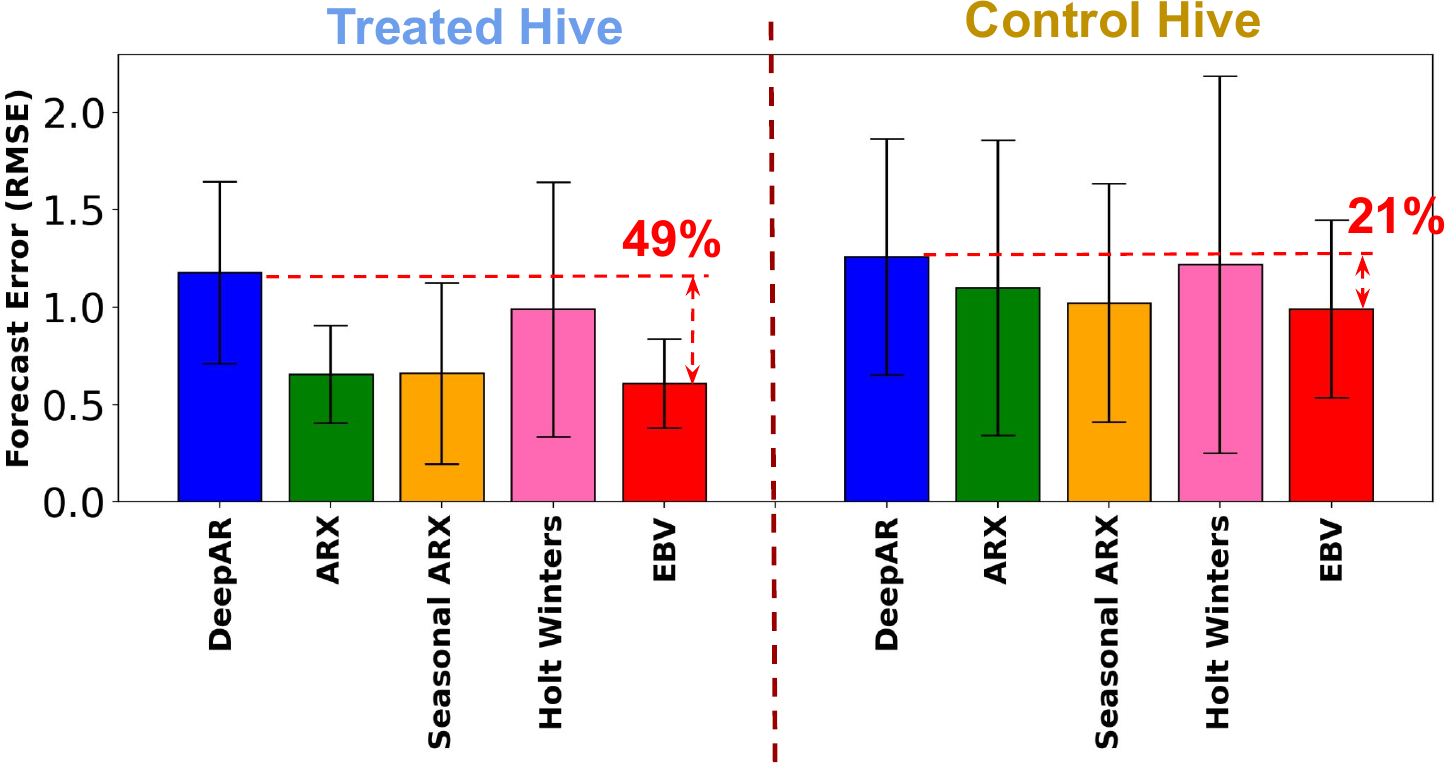}\vspace{-5pt}
\caption{\myTag{\method wins} against baseline in terms of forecasting accuracy (up to 49\% improvement). Error bars show 1 standard deviation.}
\label{fig:accuracy}
\end{figure}

\smallskip\noindent\textbf{Accuracy.}  Figure \ref{fig:accuracy} shows the comparison of the average RMSE of all sequences among baselines and \method for different hive types. \method shows an average improvement of up to 49\% over baselines in terms of accuracy. The average RMSEs for all experimental hives are also reported in Figure \ref{fig:work_flow}(c). Each point corresponds to a (hive-id, forecasting method) pair. We show that all points (except a few) fall in the upper left region of the red line (average RMSEs of \method), indicating better forecasting accuracy of \method than baselines.

\begin{figure*}
    %\subfloat[Control Hive (fitted using \methodOne)]{
    %    \includegraphics[height = 3.7cm, width = 8.5cm]{Figures/model_1_265.pdf}
    %}
    \subfloat[Control Hive (High Heat Stress)]{
        \includegraphics[height = 4cm]{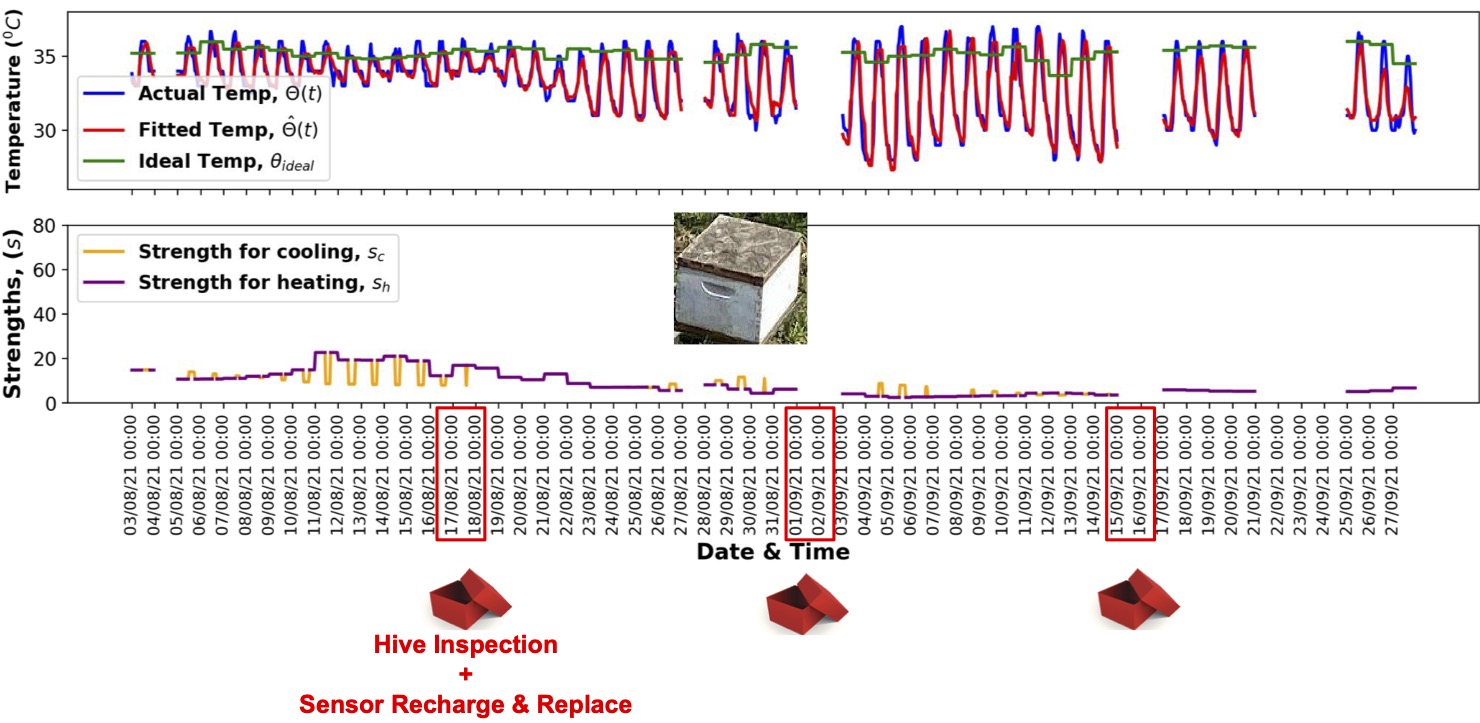}
    }%\\\vspace{10pt}
    %\subfloat[Treated Hive (fitted using \methodOne)]{
    %    \includegraphics[height = 3.7cm, width = 8.5cm]{Figures/model_1_260.pdf}
    %}
    \subfloat[Treated Hive (Low Heat Stress)]{
        \includegraphics[height = 4cm]{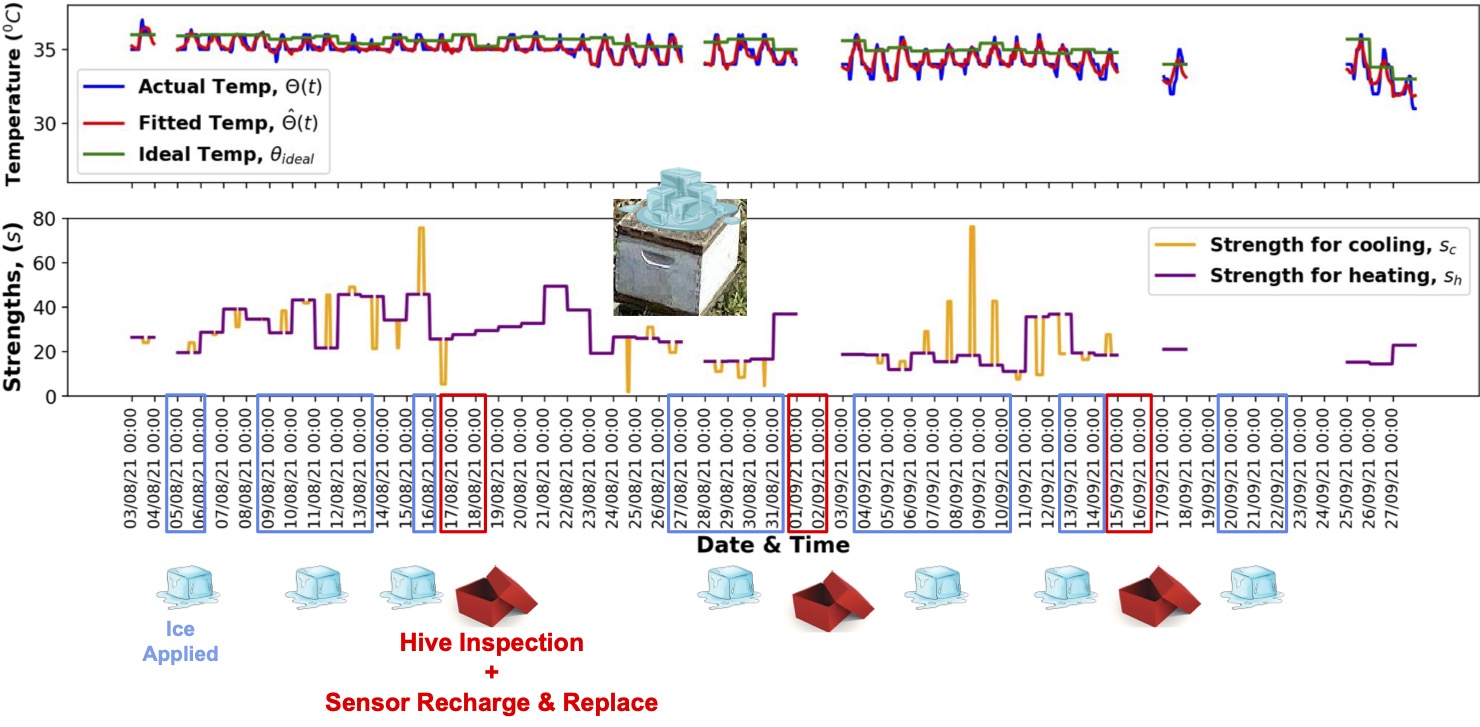}
    }\vspace{-5pt}
    \caption{\myTag{\method captures the effect of ice treatment:} Bees in treated hives are stronger than the ones in control hives. }
    \label{fig:curveFitting}
\end{figure*}

ARX and seasonal ARX do good forecasting (second-best), but our proposed method, \method has an additional advantage over any other method: parameters of baselines are not explainable whereas, \method can better explain the bee strengths with the parameters $\strengthCooling$ and $\strengthHeating$. Also, the intuition of \method conforms to the domain experts' expectations: increased fluctuation in core temperature $\thetaCore$ indicates decreased bee strengths ($\strengthCooling, \strengthHeating$) to maintain $\thetaCore$ and deteriorating hive health (Figure \ref{fig:curveFitting}).

\subsection{Q2 - Explainability} \label{sec:explainability}

As mentioned in Section ~\ref{sec:eval_segmentation}, each cutpoint and the change in parameters between adjacent segments (Figure \ref{fig:segmentation}%and \ref{fig:segmentation_2}
) are indications of events affecting thermoregulation and hive health. 

\smallskip\noindent\textbf{Effect of intensive hive inspection.} According to domain experts, the effect of intense hive inspections is reflected through the change in $\thetaCore$ within a few days following the inspection. %Below we give a detailed explanation.

%\begin{tight_itemize}
%\item Intensive Inspection (2021 Data): 
For the control hive (Figure \ref{fig:segmentation}(a)), $\strengthCooling$ went up by 38\% while the $\strengthHeating$ went down by 35\% after the first cut-point. It indicates bees’ struggle to manage the cleanliness and $\thetaCore$ following hive inspection and sensor replacement. After the second cut-point, both $\strengthCooling$ and $\strengthHeating$ decreased drastically by 82\% and 55\%, indicating severely affected thermoregulation and weak bees.
%\item Regular Inspection (2023 Data): No discontinuity detected in Figure \ref{fig:segmentation_2} is related to hive inspections. Rather they can be explained by events reported during inspections: capped queen cells followed by hive absconding.

%\end{tight_itemize}

%\smallskip\noindent\textbf{Effect of harsh hive inspection.} The change in parameters is more frequent and drastic in the control hive (more cut points) than in the treated one (fewer cut points). Below we give a more detailed explanation.

%For the control hive (Fig.\ \ref{fig:segmentation}(a)), $\strengthCooling$ went up by 38\% while the $\strengthHeating$ went down by 35\% after the first cut-point. It indicates bees’ struggle to control the cleanliness and $\thetaCore$ following hive inspection and sensor replacement. After the second cut-point, both $\strengthCooling$ and $\strengthHeating$ decreased drastically by 82\% and 55\%, indicating severely affected thermoregulation and weak bees.

%For the treated hive in Fig.\ \ref{fig:segmentation}(b), \myAlgorithm detects only one cut-point (only after the first inspection) with a reduction of 40\% and 48\% in $\strengthCooling$ and $\strengthHeating$ respectively.  No parameter changes were detected for the following two events. This indicates that the bees were healthy enough to combat the effect of harsh hive inspections and control the fluctuation in $\thetaCore$. 

\smallskip\noindent\textbf{Effect of ice treatment on thermoregulation.}
Figure \ref{fig:curveFitting} shows the reconstructed curves (red line) for the recorded $\thetaCore$ (blue line) assuming daily fluctuation in parameters, $\thetaIdeal$ (green line: {\em top graphs}) and bee strengths, $\strengthCooling$, $\strengthHeating$ (orange and purple lines: {\em bottom graphs}) over the experiment period.
%It is evident that the blue and red line aligns well %and the fitting is more accurate 
%for \myTag{\methodTwo}. 
%The second part of each graph %in Figure \ref{fig:curveFitting} is the fluctuation in the strengths of bees over the experimental period. 

In the bottom graphs, we can observe (i) the low strength of bees (in the range of [3,25]; Figure \ref{fig:curveFitting}(a)) in control hives compared to treated ones (mostly in the range of [20,80]; Figure \ref{fig:curveFitting}(b)), (ii) comparatively drastic change in bee strengths ($\strengthCooling, \strengthHeating$) and increased fluctuation in $\thetaCore$ of the control hive, following intensive inspection.

For the treated hive in Figure \ref{fig:segmentation}(b), \myAlgorithm detects only one cut-point (only after the first inspection) with a reduction of 40\%
and 48\% in $\strengthCooling$ and $\strengthHeating$ respectively.  No parameter changes were detected for the following
two events. This indicates that the bees were healthy enough to combat the aftereffect of intensive inspections and control the fluctuation in $\thetaCore$. 

These results coincide with domain experts' expectations: ice packs during hot days will help bees regulate $\thetaCore$ and manage the aftereffects of intensive inspections.

%\smallskip\noindent\textbf{Motivation to thermoregulate.} Broods and larvae are the motivation for bees to tightly maintain $\thetaCore$. In Figure \ref{fig:segmentation_2}, a sudden decrease in $\strengthHeating$ by 91\% after the first cut indicates bees' reluctance to thermoregulate due to no developing brood following the emergence of a queen. 

%The values of $\strengthCooling (\neq 0)$ and $\strengthHeating (\approx 0)$ after the second and third cut-points coincide with the fact that robber bees and the leftover resources cause $\thetaCore$ and $\thetaExt$ not become same even after absconding.

%\smallskip\noindent\textbf{Effect of mite infestation.} No change in $\strengthHeating$, $\strengthCooling$ till 22 August indicates that bees are resilient and tightly managing $\thetaCore$. A sudden decrease of 100\% and 95\% in $\strengthCooling$ and $\strengthHeating$ on 23 August (second cut-point) indicates exhausted and weak bees. It coincides with the report that they were aggressive on 24 August and collapsed the next day. Domain experts believe that bees were too weak to handle the effect of hive opening.

\begin{figure}
\centering
    %\subfloat[Pdf of $\strength$ vs $\strength$\vspace{-5pt}]{
    %    \includegraphics[height = 4cm, width = 7cm]{Figures/w_count.pdf}
    %}\hspace{2cm}
    %\subfloat[$\strengthCooling$ vs $\strengthHeating$ in logarithmic scale\vspace{-5pt}]{
        \includegraphics[height = 4cm]{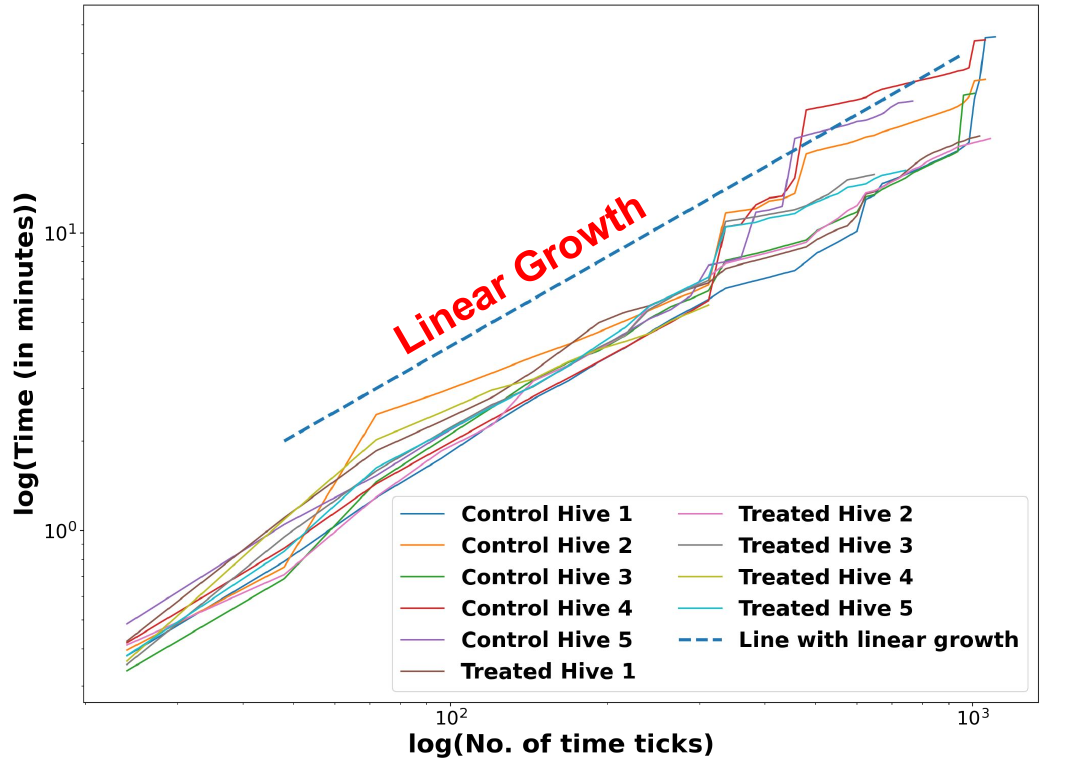}\vspace{-5pt}
    %}
\caption{\myTag{\method is linear} with input size : growth parallel to diagonal.} 
\label{fig:linear_time}
\end{figure}

\subsection{Q3 - Scalability}\label{sec:scalability}

Our proposed model, \methodTwo is fast, 
% scalable, 
and takes only about 20 minutes on average to process and fit two months of data for a well-defined search space on the laptop described above. %It is also linear on input size (Figure \ref{fig:linear_time}).
% except for some sequences where the optimizer takes more time than average to converge and find a global minimum. 
\methodTwo is easily parallelizable as the data from each day and each hive is independent.
% For a well-defined search space, it takes equal time to process every temperature data sequence. 
We conclude with the following lemmas.
\begin{lemma}
     Reconstruction using \methodTwo is linear in time, i.e., for $t$ time ticks, the time complexity of reconstruction is O(t).
\end{lemma}

\begin{proof}
     We use Levenberg-Marquardt (LM) optimization, and the time for convergence is linear with the number of time ticks. $\quad\blacksquare$ 
 \end{proof}

%Each line (except the line of slope 1) in Figure ~\ref{fig:linear_time} demonstrates the time to reconstruct data for a particular experimental hive. 
All (solid) lines in Figure ~\ref{fig:linear_time} (time to reconstruct two months of data for experimental hives in logarithmic scale) are parallel to the reference line of linear growth (dashed blue line), indicating scalability.

% \begin{proof} Straightforward - omitted for brevity.
% \end{proof}

\begin{lemma}
    Segmentation using \myAlgorithm is linear in time, i.e., the time complexity for finding m segments in a given dataset $\mathcal{D}$ 
    with t time ticks is $O(t \times m)$. 
\end{lemma}

\begin{proof}
For a time sequence with $t$ time ticks, we use a greedy algorithm
to find the $m$ cut-points.
Each cut-point requires one scan of the sequence $O(m)$, and thus
the complexity is $O(t \times m)$.
$\quad\blacksquare$
\end{proof}

\subsection{Q4 - Observations} \label{sec:observation}
\textbf{Effect of intensive hive inspection.} 
As described in Section \ref{sec:eval_segmentation}, our algorithm \myAlgorithm detected 80\% more cut-points in total for control hives than for treated ones for the same number (3 per hive) and type of (intensive) inspection.
That is: 
% Hence, we conclude with the following observation.

%\vspace{-3pt}
\begin{observation} 
Control (= untreated) hives suffer more from intensive hive inspections.
\end{observation}
%\vspace{-6pt}
That is, control hives show more discontinuities and 
less control over their temperature. %\myAlgorithm therefore allows beekeepers to quantify the impact of their management on bee health.
% More cut-points are detected in the control hive data compared to the data from the treated one. The more the no. of cut-points, the more frequently the parameters (i.e., bee strength) change. 

\begin{figure}
\centering
    %\subfloat[Pdf of $\strength$ vs $\strength$\vspace{-5pt}]{
    %    \includegraphics[height = 4cm, width = 7cm]{Figures/w_count.pdf}
    %}\hspace{2cm}
    %\subfloat[$\strengthCooling$ vs $\strengthHeating$ in logarithmic scale\vspace{-5pt}]{
        \includegraphics[height = 4cm]{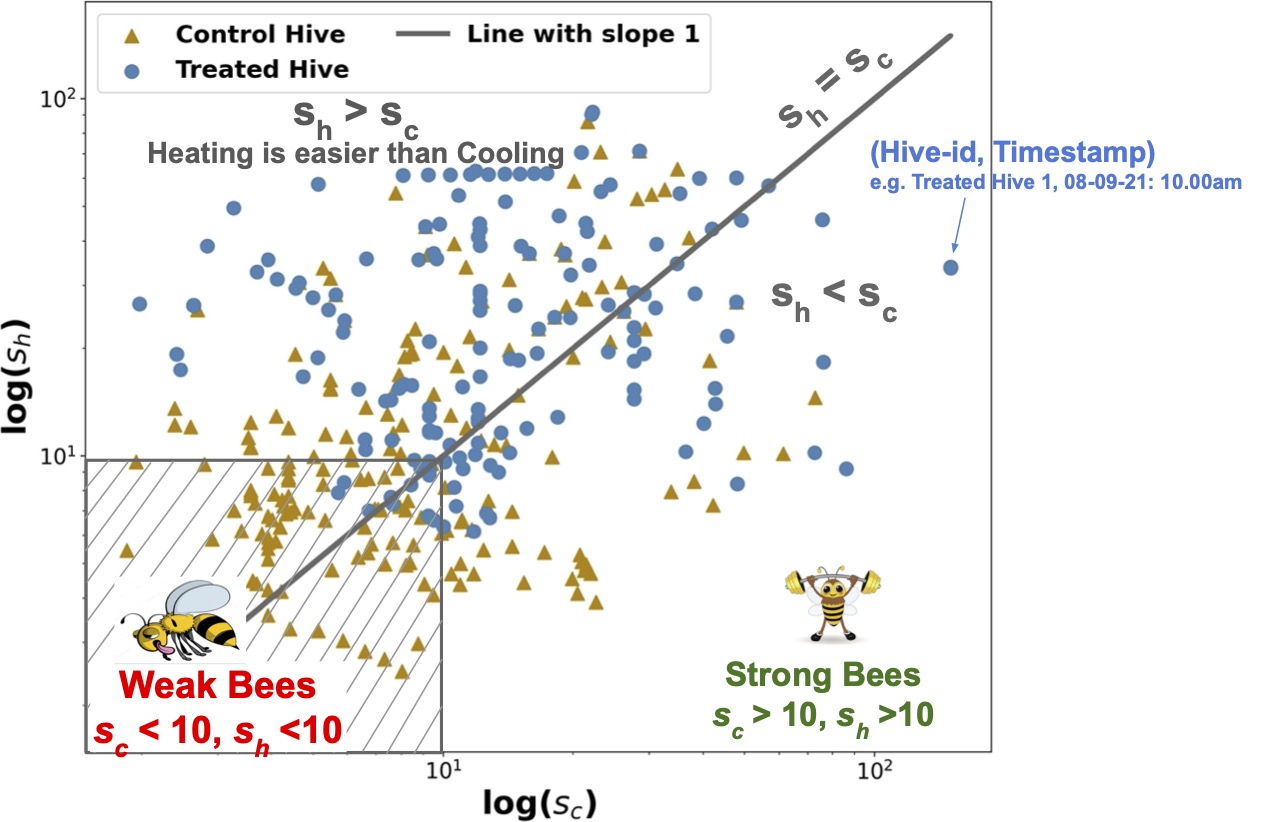}%\vspace{-5pt}
    %}
\caption{\myTag{\method spot differences}: heating/cooling strength $\strengthCooling$ vs $\strengthHeating$ in logarithmic scale. It corroborates that bees in treated hives are stronger.}
\label{fig:w_count}
\end{figure}

\smallskip\noindent\textbf{Effect of ice treatment on thermoregulation.}  We demonstrate the distribution of bee strengths over the experimental period from different hive settings ({\em 2021 Data}) as a logarithmic scatter plot of  $\strengthCooling$ vs $\strengthHeating$ 
(cooling-strength vs heating-strength) in Figure \ref{fig:w_count}. 
Every data point corresponds to a (hive-id, timestamp) pair.

Notice that bees in control (= untreated) hives show a lower ability to heat and cool their hives.
%have the weakest cooling and heating abilities.
\hide{
We can see that more data points from control hives are in the weaker region $(\strengthCooling < 10, \strengthHeating <10)$ than the treated ones. Most of the data points from treated hives are in the stronger region representing stronger bees and better thermoregulation activities. Therefore, we make the following observation.
}
%\vspace{-6pt}
\begin{observation} Bees in treated hives are stronger, i.e., they have better thermoregulation ability than the ones in control hives.
%(see Figures \ref{fig:curveFitting} \& \ref{fig:w_count}).
\end{observation}

\smallskip\noindent\textbf{Difference in strength during heating and cooling.} The gray line in Figure \ref{fig:w_count} divides the graph into two regions: (i) upper left: heating is easier than cooling (ii) lower right: cooling is easier than heating. 
Notice that most data points fall into the upper left region, coinciding with the domain expert's expectation:
%\vspace{-6pt}
\begin{observation} Heating a hive is easier than cooling it down.
\end{observation}

%\smallskip\noindent\textbf{Motivation to thermoregulate.} As shown in Figure \ref{fig:segmentation_2}, sudden increased fluctuation of $\thetaCore$ and drastic drop in strengths ($\strengthHeating$) a few days after the emergence of queen coincides with experts' opinion:
%\vspace{-6pt}
%\begin{observation}
    %No broods and larvae in the colony, no need to tightly maintain the hive's core temperature. 
%    In the absence of developing broods in the colony, bees don't need to tightly regulate the hive’s core temperature.
%\end{observation}

\section{Conclusions, Significance and Impact}\label{sec:conclusion}
We proposed  \method (Electronic Bee-Veterinarian)  to help beekeepers monitor and analyze the health of their hives and take preventive actions.
% \method is {\em principled} (based on the diffusion equstion), automatic scalable, etc

\method is (a) principled, using  physics (thermal diffusion equations) and control theory (feedback-loop {\em `split'} P-controllers);  %effective in analyzing temperature sensor data from hives, and 
(b) interpretable by beekeepers with only a few parameters (i.e., bee strengths for heating and cooling); (c) effective in fitting, forecasting, and detecting events affecting hive health; (d) scalable; and (e) informative, agreeing with experts' expectations.
% \method was
%consists of the model \methodTwo and the algorithm \myAlgorithm, and they were tested on multiple real-world time series data and 
%found to be accurate in fitting, forecasting, and segmentation. The discontinuities detected by \method coincide with the external interventions and events that could stress the hive, indicating the practical validity of the approach. 
% The proposed EBV provides a substantial improvement over prior work at hive monitoring that lacked explainability and strong evidence of effectiveness, and it will thus help address the need for functional bee health monitoring systems. 

\smallskip\noindent{\bf Significance.}
Honeybees are vital for pollination of crops. Novel monitoring tools to safeguard bees are a matter of global urgency. Our proposed model, \method will be of broad interest to beekeepers, growers, and researchers studying all facets of pollinator health all over the world. A reliable model with confirmed analytic capabilities will be a game changer for the global beekeeping industry. 

\smallskip\noindent\textbf{Impact.} A preliminary version of our model received positive feedback from local beekeepers at the UC Riverside Bee Health conference in 2022~\cite{bee_conference}. We are in constant contact with them and co-organizing yearly events with CIBER to keep them updated about our endeavor. 

\smallskip\noindent{\bf Reproducibility.} As mentioned in Section \ref{sec:intro}, our code is on GitHub:
\url{https://github.com/rtenlab/EBeeVet}; 
and our collected dataset is available upon request.

\section{Acknowledgments}
This work was partially supported by a UC Multicampus Research Programs and Initiatives (MRPI) award (\# M21PR2306). We extend special thanks to Jamison Scholer for his help in managing the bee hives and Mohsen Karimi for his help in collecting the data.

\end{document}